\documentclass{ieeeaccess}

\usepackage{algorithmic}
\usepackage{fontawesome}
\usepackage{adjustbox}
\usepackage{graphics}
\usepackage{amsfonts}
\usepackage{graphicx}
\usepackage{textcomp}
\usepackage{placeins}
\usepackage{amsmath}
\usepackage{amssymb}
\usepackage{cite}
\usepackage{url}
\allowdisplaybreaks

\def\BibTeX{{\rm B\kern-.05em{\sc i\kern-.025em b}\kern-.08em T\kern-.1667em\lower.7ex\hbox{E}\kern-.125emX}}

\DeclareMathOperator*{\minimize}{\textbf{\textit{minimize~}}}
\DeclareMathAlphabet{\mathcal}{OMS}{cmsy}{m}{n}

\newcommand{\mymtx}[1]{\mathbf{\MakeUppercase #1}}
\newcommand{\myvec}[1]{\mathbf{\MakeLowercase #1}}

\begin{document}
% \history{Date of publication xxxx 00, 0000, date of current version xxxx 00, 0000.}
\doi{ |~~Preprint~July,~2022}%10.1109/ACCESS.20xx.DOI}

\title{Unfolding AIS transmission behavior for vessel movement modeling on noisy data leveraging machine learning}

\author{
    \uppercase{Gabriel Spadon},\hspace{-.8mm}\authorrefmark{$1\star$}
    \uppercase{Martha D. Ferreira},\hspace{-.8mm}\authorrefmark{$1\star$}
    \uppercase{Amilcar Soares},\hspace{-.8mm}\authorrefmark{$2$}
    \uppercase{Stan Matwin}\hspace{.1mm}\authorrefmark{$1,3\dagger$}
    \vspace{.1cm}  % bug-fix
}
\address[$1$]{Institute for Big Data Analytics --- Dalhousie University, Halifax - NS -- Canada\\[.05cm]}
\address[$2$]{Department of Computer Science --- Memorial University of Newfoundland, St. John's - NL -- Canada\\[.05cm]}
\address[$3$]{Institute of Computer Science --- Polish Academy of Sciences, Warsaw -- Poland}

% \tfootnote{$\star$ Authors have contributed equally to this work.}
\tfootnote{$\dagger$ Corresponding author (e-mail: stan@cs.dal.ca).\\[.05cm]
$\star$ Authors have contributed equally to this work.}

% \markboth{{Spadon \headeretal} | Modeling collective AIS transmissions behavior for forecasting vessel movement on irregular timing data}

% \corresp{Corresponding author: Stan Matwin (e-mail: stan@cs.dal.ca).}

\begin{abstract}
    The oceans are a source of an impressive mixture of complex data that could be used to uncover relationships yet to be discovered.
    Such data comes from the oceans and their surface, such as Automatic Identification System (AIS) messages used for tracking vessels' trajectories.
    AIS messages are transmitted over radio or satellite at ideally periodic time intervals but vary irregularly over time.
    As such, this paper aims to model the AIS message transmission behavior through neural networks for forecasting upcoming AIS messages' content from multiple vessels, particularly in a simultaneous approach despite messages' temporal irregularities as outliers.
    We present a set of experiments comprising multiple algorithms for forecasting tasks with horizon sizes of varying lengths.
    Deep learning models ({\it e.g.}, neural networks) revealed themselves to adequately preserve vessels' spatial awareness regardless of temporal irregularity.
    We show how convolutional layers, feed-forward networks, and recurrent neural networks can improve such tasks by working together.
    Experimenting with short, medium, and large-sized sequences of messages, our model achieved $36/37/38\%$ of the Relative Percentage Difference -- the lower, the better, whereas we observed $92/45/96\%$ on the Elman's RNN, $51/52/40\%$ on the GRU, and $129/98/61\%$ on the LSTM.
    These results support our model as a driver for improving the prediction of vessel routes when analyzing multiple vessels of diverging types simultaneously under temporally noise data.
\end{abstract}


\titlepgskip=-15pt

\maketitle

\section{Introduction}
\label{sect:introduction}

\PARstart{O}{ver} the years, we have been experiencing a massive maritime vessel trajectory network\footnote{Referred in this work as a complex network, network, or graph.} expansion powered by globalization and the evolution of transportation~\cite{Carlini2021:MaritimeNetworks}.
Maritime navigation is essential in passenger transportation, tourism, and fishing~\cite{Massimiliano2021:HumanMobility, Millefiori2021:MaritimeMobility, Ribeiro2020:TrajectoryAnalytics, Walker2019:MarineTransportationEffects}.
In addition, it has been historically used for trading between territories and countries worldwide~\cite{Ducruet2020:MaritimeNetworksReview, Wiegmans2020:HistoricalFreight, Gastner2014:TailedTraffic}.
Over the centuries, many efforts have been focused on forecasting wind, waves, and weather to be prepared for non-ideal navigation conditions~\cite{Wang2021:WindForecasting, Kourafalou2015:OceanForecasting, Papageorgiou2016:MarineTourism}.
However, ocean activities are far from controllable.
In addition to climate-related risks, there exist significant concerns of piracy ({\it e.g.}, armed robbery and hijackings), equipment defects, and ship collisions, among others~\cite{Marchione2013:MaritimePiracy, Goerlandt2011:ShipCollision, Posada2011:Piracy, Chen2019:ShipCollision, petry2020challenges}.

A convenient way of preventing or responding to adverse events found in the sea is tracking vessels' trajectories through Automatic Identification System (AIS) messages~\cite{Robards2016:AIS-Review}, which are part of a more extensive system that monitors maritime navigation activity~\cite{Soares2019:CRISIS}.
These messages are transmitted over radio or satellite at ideally periodic time intervals~\cite{Yang2019:AIS-BigData}, containing information on vessel identification and current status.
Details such as geographical coordinates, course, and speed over the ground are also included~\cite{Harati-Mokhtari2007:AIS-Reliability}, turning AIS into a supporting technology for vessel tracking with acknowledged relevance to ocean monitoring~\cite{Norris2007:AIS, Lee2019:AISMaturity}.

The literature on transportation systems has been leveraging the volume AIS data and its overly sequential nature to develop a range of vessel trajectory forecasting techniques~\cite{Nguyen2018:MaritimeSurveillance, Patmanidis2016:RouteEstimation, Zhang2018:TrajectoryDelimiter, Uney2019:StochasticGenerativeModels}.
The interest in forecasting trajectories comes from the capability to estimate vessel routes, which increases the safety and reliability of marine transportation~\cite{FAGHIHROOHI2014363} and enhances oceans' situational awareness~\cite{le2013unsupervised}.
Popular techniques employed to address those tasks are based on Recurrent Neural Networks (RNNs)~\cite{Wang2020:BiGRU, Park2021:BiLSTM, Han2019:GRU}, Auto-Encoders (AEs)~\cite{Murray2021:DLFramework, Murray2020:DualAutoencoder, Capobianco2021:AutoEncoder}, and Convolution Neural Networks (CNNs)~\cite{Chen2020:TCNN}.
More recent techniques have focused on Graph Neural Networks (GNNs)~\cite{eljabu2021anomaly} and Network Embeddings~\cite{nguyen2018multi}.
Several techniques have focused on the impact of multi-directional~\cite{bao2022improved} and multi-layer~\cite{liu2020hybrid, sagheer2019unsupervised} RNNs for enhancing forecasting tasks in a regression fashion.
Others have opened the discussion on leveraging multiple trajectories to improve trajectory modeling and mobility pattern understanding~\cite{wang2017vessel}.
However, the related literature still lacks investigation when these scenarios merge and are composed of streaming data.
These conditions are relevant to time-sensitive tasks requiring near real-time inference abilities, which are challenging due to a lack of extensive data preprocessing possibilities to deal with outliers.

Trajectories' irregular timing is typically caused by transmission delays, lack of signal coverage, equipment defects, and interference despite vessels sending AIS messages periodically ({\it i.e.}, every few seconds or minutes).
The irregular timing is a preprocessing drawback to overcome when working with AIS data because it can bring inconsistency in picturing a clear vessel route, jeopardizing the maritime domain awareness~\cite{Tetreault2005:AIS-MDA}.
Moreover, in some cases, such behavior might be deliberate and related to irregular maritime activities~\cite{dAfflisio2021:MaliciousSpoofing}, but those are usually exceptions among a population of AIS messages.
Notice that such irregularity is tied to the vessel's AIS transceiver technology and whether the message will be captured by low-range radio or long-range satellite receivers.
Vessels near the shore are usually captured by radio and far away by satellites.
Working in a large geographical area, one would be subject to data from multiple sources, including different transmission behaviors.
These are tied to the type and the location of the vessel transmitting and the receiver capturing the AIS message.

Previous works in the literature consistently adopted a trajectory interpolation approach to address this issue, which has been actively used as a resource for better trajectory planning and forecasting.
Such an approach inserts virtual messages in the vessel trajectory to smooth the timing irregularity, allowing the trajectory to be strictly periodic~\cite{Li2019:1DTrajectorySmoothing, Li2019:2DTrajectorySmoothing}.
Therefore, the authors transform the AIS data into a well-behaved discrete-contained time series ({\it i.e.}, an ordered sequence).
However, this approach can introduce uncertainty in vessel routes when the gap between two consecutive AIS messages is too large, which would alter the trajectory's data distribution and picture an inaccurate trajectory.
This would be the case for vessels with mobility patterns different from in-line sailing, in which the geometry of the trajectory matters ({\it e.g.}, fishing and military vessels).
Such a disadvantage might provide modeling solutions not robust to outliers.

{
Our assumption lies in accounting for multiple vessels of varied types and the multiple numerical variables within the AIS message to overcome the timing irregularity and achieve better performance on the foreseen non-preprocessed AIS data, covering larger geographical areas regardless of the AIS message ({\it i.e.}, either radio-or satellite-based) type.
Using this approach, we intend to leverage information that is usually overlooked to increase the ability of the model to learn the intricacies of space ({\it i.e.}, from where the vessel is transmitting) and time ({\it i.e.}, since the last message received was acknowledged) to increase the model's generalization capability over different trajectories and mobility patterns.
}

{
Unlike traditional trajectory forecasting, smoothing, and compressing algorithms ({\it i.e.}, series reduction), our focus is on the entire continuous-defined content ({\it e.g.}, latitude, longitude, Course over Ground -- COG, and Speed over Ground -- SOG) of the subsequent AIS message in the transmission sequence rather than being concerned only with the next coordinates of the vessel.
Therefore, we do not intend to replace traditional series reduction techniques such as the Douglas-Peucker~\cite{hershberger1992speeding}, and the same holds for Ornstein–Uhlenbeck processes~\cite{coscia2018multiple} for trajectory approximation or clustering for mobility pattern analysis.
Our proposal is to be used in cases where the AIS messages are unavailable and can be reconstructed simultaneously with other vessels in the trajectory network.
As part of the AIS forecasting task, the vessel's positioning is included, and the trajectory is preserved but not at the same level of granularity as traditional trajectory forecasting techniques.
The same holds for smoothing-based techniques because our model intends to foresee AIS transmissions.
Thus, the number of expected messages is the same as the real-world AIS transmission system ideally receives.
}

{
In this sense, this paper focuses on accurately representing the transmission system for maximizing generalization over mixed-typed vessels indistinctly.
Our goal consists of minimizing the shared error between the predicted and observed AIS messages coming from heterogeneous vessel tracking sources.
To the best of our knowledge, this approach has not yet been studied from the perspective of maritime vessel trajectories due to its inherent timing complexity and volume of data in the form of AIS messages.
It could offer a unique milestone for future research with similar patterns.
Hence, we seek a sufficiently robust model for different data distributions and outliers arising from the delta time between consecutive AIS messages.
Therefore, we propose using an artificial neural network model mixing single-dimension convolution layers, recurrent neural networks, and feed-forward neural networks into a single architecture for multi-task and multivariate AIS transmission forecasting that achieves increased performance in predicting the intermediate states of the vessel trajectory network as upcoming AIS messages.
}

{
Our results are based on extensive experiments contrasting the capability of several machine and deep learning models, which are bounded to univariate or multivariate samples.
However, the problem we are tackling requires considering multiple variables across multiple instants of time for multiple samples due to different data distributions and mobility patterns arising from different vessels.
These models were tested multiple times for different sets of samples and variables.
Our results comprehensively compare the forecasting of AIS messages for single and multiple vessels, considering one or more variables.
We cover a range of baselines driven to {\bf (A)} single trajectories with multivariate estimators, {\bf (B)} single trajectories with multiple univariate estimators, and {\bf (C)} multiple trajectories with multivariate estimators.
}

The results show that our model improves the prediction of vessel routes when simultaneously analyzing multiple vessels of diverging types.
This translates into a model that, on average, provides more accurate forecasting results over multiple trajectories rather than a model tailored for a single class of vessels or trained on long historical sequences of AIS messages of a single vessel.
Moreover, the results point out that traditional machine learning models struggle to generalize over different vessels, while deep learning models can better capture the temporal irregularity and spatial features while simultaneously describing multiple vessels' trajectories.
In such a case, deep learning models achieve improved results over competing algorithms, mainly when working with convolutional layers.
In experiments with short, medium, and large-sized AIS messages sequences, the proposed model achieved $36/37/38\%$ of the Relative Percentage Difference (RPD) -- the lower, the better, whereas we observed $92/45/96\%$ on the Elman's RNN, $51/52/40\%$ on the Gated Recurrent Unit (GRU), and $129/98/61\%$ on the Long-Short Memory (LSTM) network.
In addition to the performance improvement derived from our alternative network architecture, we also observed that our model was more numerically stable over the various experiments using different window and horizon sizes, showing better performance in forecasting short and long AIS message sequences for multiple vessels.
Contrarily, other models revealed varying performance over different-sized AIS message sequences.

In conclusion, our contributions can be summarized as:
\begin{itemize}
    \item A new perspective for AIS transmission behavior modeling accounting for the full continuous-valued content of the AIS message under temporal noise effect;
    \item A comprehensive benchmark with several machine and deep learning models submitted to the same forecasting task on horizon sizes of comprehensive lengths;
    \item A methodological pipeline that describes how to capture the multiple data distributions on the temporal data for different vessel trajectories in a single model; and,
    \item A proposed model based on recurrent neural networks, convolution, and feed-forward layers to achieve increased performance regardless of the vessel type.
\end{itemize}

This article is organized into three sections apart from the Introduction in Section~\ref{sect:introduction}.
Section~\ref{sect:methodology} states the problem, describes the dataset, and presents the methodology.
Section~\ref{sect:results-discussion} review the main results and discusses our findings.
Section~\ref{sect:conclusions} addresses the conclusions and future works.
The supplementary material includes details on the baseline experiments.

\section{Methodology}
\label{sect:methodology}

\subsection{Problem Formalization}
\label{sect:problem-formalization}

AIS messages contain different static and dynamic information describing vessel trajectories that vary according to the different ocean and traffic monitoring applications in which they are used.
In this paper, we defined an AIS message of a vessel as an event $v$, which is defined as $\vec{v} = \left<\rho, \omega, \psi, \epsilon, \mu\right>$, having latitude $\rho$, longitude $\omega$, time $\psi$, Course Over Ground (COG) $\epsilon$ and Speed Over Ground (SOG) $\mu$ as attributes.

The sequence of AIS messages of a vessel shapes its trajectory, which has a non-standard ({\it i.e.}, varying) length.
Thus, we define the trajectory of a vessel as $\tau_i = \{V_{\tau_i}, E_{\tau_i}\}$, being a sequence of ordered events $v \in V_{\tau_i}$ connected by an edge $e \in E_{\tau_i}$.
The edges are unweighted in our formulation, but they could represent, the distance $\mathcal{D}$ between the source $\vec{v}_n$ and target $\vec{v}_{n+1}$ AIS messages in a sequence, such that $e = \left<\vec{v}_n, \vec{v}_{n+1}, \mathcal{D}_{n, n+1}\right>,~ \forall n \leq |V_{\tau_i}| - 1$.

Through such data, it is possible to derive a disconnected graph $T$ by modeling the dataset's vessel trajectories as components.
$T = \left\{\tau_0, \tau_1, \ldots, \tau_c \right\}$ is a network of multiple connected components, in which $\tau_i \in T~\forall~0 \leq i \leq c$ and $c$ is the total number of different vessels.
The trajectories are not segmented\footnote{In this work, trajectories and vessels are treated as the same.}, so each vessel has only one sequence of AIS messages that varies according to the number of messages transmitted by the vessel and received by radio or satellite receivers.
Knowing that different vessels cannot occupy the same space at the same time, $T$ is under the condition that $V_{\tau_i} \cap V_{\tau_j} = \emptyset \land E_{\tau_i} \cap E_{\tau_j} = \emptyset,~ \forall \left<i,j\right> \leq |T|,~ i \neq j$.

In terms of sequences and series, each trajectory $\tau \in T$ is composed of a sequence of ordered events $V_{\tau} = \left<\vec{v}_0, \vec{v}_1, \ldots, \vec{v}_p\right>$, where $p \in \mathbb{N}_{+}$ is the total number of events which varies for each vessel.
The events are sets of spatiotemporal features describing the vessel trajectory information at different instants of time, such as given by\\ $V_{\tau} = \left<\left<\rho, \omega, \psi, \epsilon, \mu\right>_0, \left<\rho, \omega, \psi, \epsilon, \mu\right>_1, \ldots, \left<\rho, \omega, \psi, \epsilon, \mu\right>_p\right>$.

In this case, the problem for a single vessel can be defined as $f : x \subset V_\tau,~ x \in \mathbb{R}_{+} \rightarrow \hat{y} \in \mathbb{R}$ and reduced to $f(x) \approx \hat{y}$, where $f$ is the network reconstruction model that given a set $x$ of observations will yield $\hat{y}$ that resembles $y$ the most, which refers to the future states of the trajectory.
Accordingly, given an arbitrary optimization function $g: \mathbb{R}^2 \rightarrow \mathbb{R}_{+}$ computed between sets $y$ and $\hat{y}$, in which $g\left(\hat{y}, y\right) \in \mathbb{R}_{+}$ and $\hat{y} \approx y$, we seek a model $f$ that minimizes $g$ for any $x \subset V_\tau$.
Notice that $x$ and $y$ are contiguously contained in the series, but that does not mean that time between AIS messages is monotonically defined.
That is because of the different types of noise faced by transmitters and receivers (see Section~\ref{sect:introduction}).

For network modeling purposes, forecasting upcoming AIS messages based on historical AIS data for an arbitrary trajectory is unfeasible when using timestamps $\psi$ as it follows a discrete probability distribution while other features are continuously defined.
When including $\Delta\mathrm{T} \in \mathbb{R}_{+}$, {\it i.e.}, the elapsed time since the last message, instead of timestamp $\psi$, the problem becomes feasible because the elapsed time has a continuous probability distribution.
Thus, we have $V_\tau = \left\{\left<\rho, \omega, \Delta\mathrm{T}, \epsilon, \mu\right>_0, \ldots, \left<\rho, \omega, \Delta\mathrm{T}, \epsilon, \mu\right>_p\right\}$, $p \in \mathbb{N}_{+}$.

In such a scenario, the relationship between time events and delta time of a trajectory $V_\tau$ is given by $\psi_{i} - \psi_{j} = \Delta\mathrm{T}_{ij},~ \forall\left<i,j\right>\leq|T|,~ i \neq j$ and $\psi_i + \Delta\mathrm{T}_{ij} = \psi_j,~ \forall\left<i,j\right>\leq|T|,~ i \neq j$, which means a timestamp can be safely inferred when at least one delta time prior in the sequence is known.

Motivated by the sequential nature of vessel trajectories, we aim to go further with the trajectory modeling problem by reconstructing the graph's topological structure and the features underneath it.
In the case of vessel trajectories, the topology and features are deeply interconnected due to the spatiotemporal nature of the AIS messages.
In such a scenario, the problem behaves non-stochastically, where the state of network node as $\vec{v}^{~t}$ depends on a sequence of $w \in \mathbb{N}_{+}$ past events $\vec{v}^{~t} = \alpha_0 \vec{v}^{~t-1} + \alpha_1 \vec{v}^{~t-2} + \ldots + \alpha_w \vec{v}^{~t-w}$ subject to a set of scaling parameters $\vec{\alpha}$.
We can define the previous relationship in terms of subsets $\bar{x} = \left\{\left<\rho, \omega, \Delta\mathrm{T}, \epsilon, \mu\right>_0, \ldots, \left<\rho, \omega, \Delta\mathrm{T}, \epsilon, \mu\right>_w\right\}$, $\bar{x} \subset \tau$ and $\bar{y} = \left\{\left<\rho, \omega, \Delta\mathrm{T}, \epsilon, \mu\right>_{w+1}, \ldots, \left<\rho, \omega, \Delta\mathrm{T}, \epsilon, \mu\right>_{w+s}\right\}$, $\bar{y} \subset \tau$, in which $w \in \mathbb{N}_{+}$ is the window of past observations and $s \in \mathbb{N}_{+}$ is the horizon to be predicted, subject to $w + s \leq |\tau|$.

We now seek a function $h$ that given $\bar{x}$ will approximate $\bar{y}$, which can be written as $h:{\mathbb{R}^{|\bar{x}|}} \to \mathbb{R}^{|\bar{y}|}$.
In such a case, $h$ represents a function that better describes a trajectory network for any vessel or subset of vessels in the dataset, capable of picturing the inner states of the vessel trajectory network in the form of foreseen AIS messages transmissions.

\begin{figure}[t!]
    \centering
    \includegraphics[width=\linewidth, height=5cm]{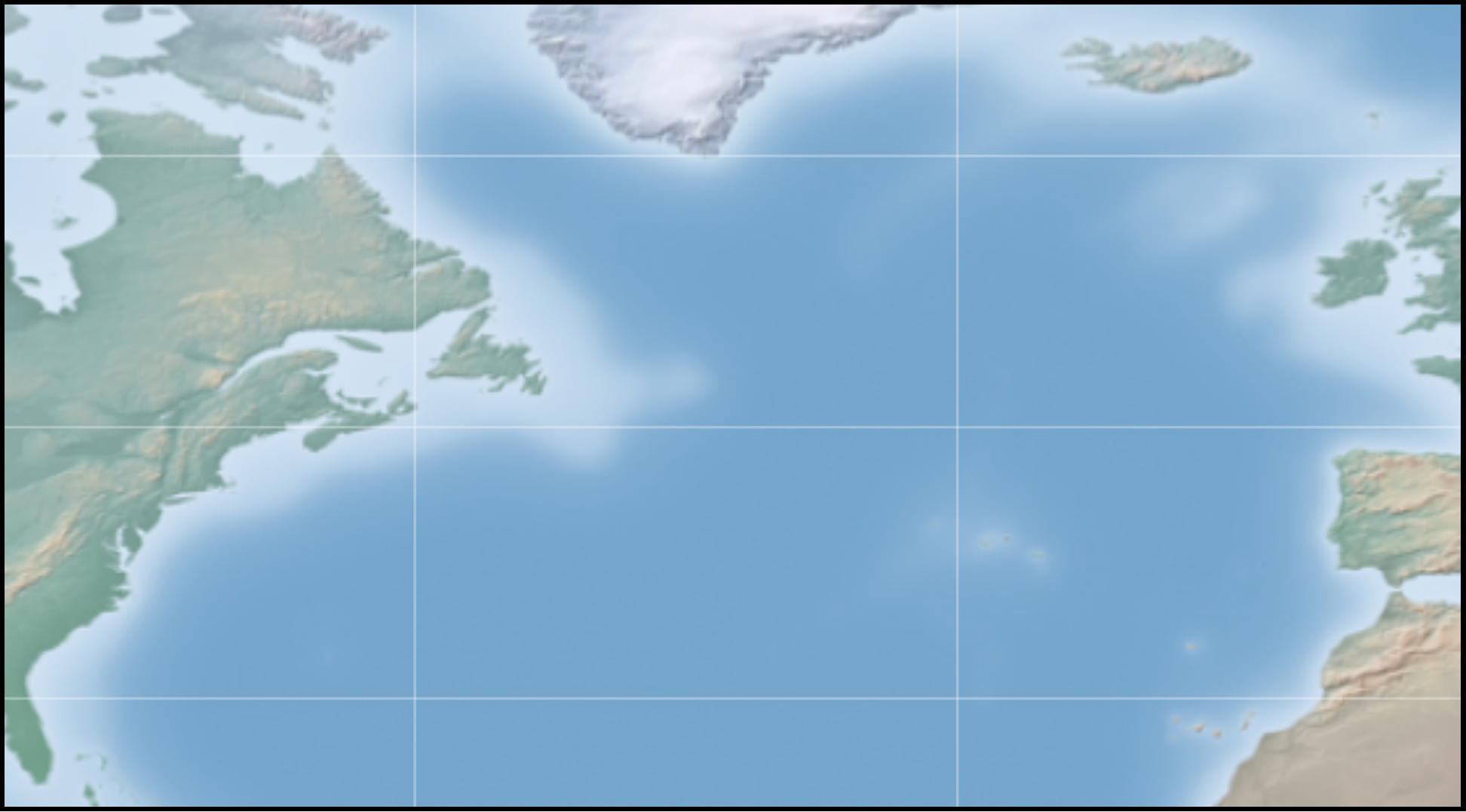}
    \caption{A cylindrical-projected map depicting the region that comprises every trajectory in the dataset of AIS messages.
    The region is a bounding-box from coordinates 23\textdegree52'14.8"\textbf{N} 82\textdegree46'58.2"\textbf{W} and 68\textdegree30'02.8"\textbf{N} 2\textdegree01'18.4"\textbf{W}.
    The trajectories in the dataset were collected between March and July 2020.}
    \label{fig:ais-map}
\end{figure}

\subsection{Spatial Coverage}
\label{sect:spatial-coverage}

\begin{figure}[t!]
    \centering
    \includegraphics[width=\linewidth, height=5cm]{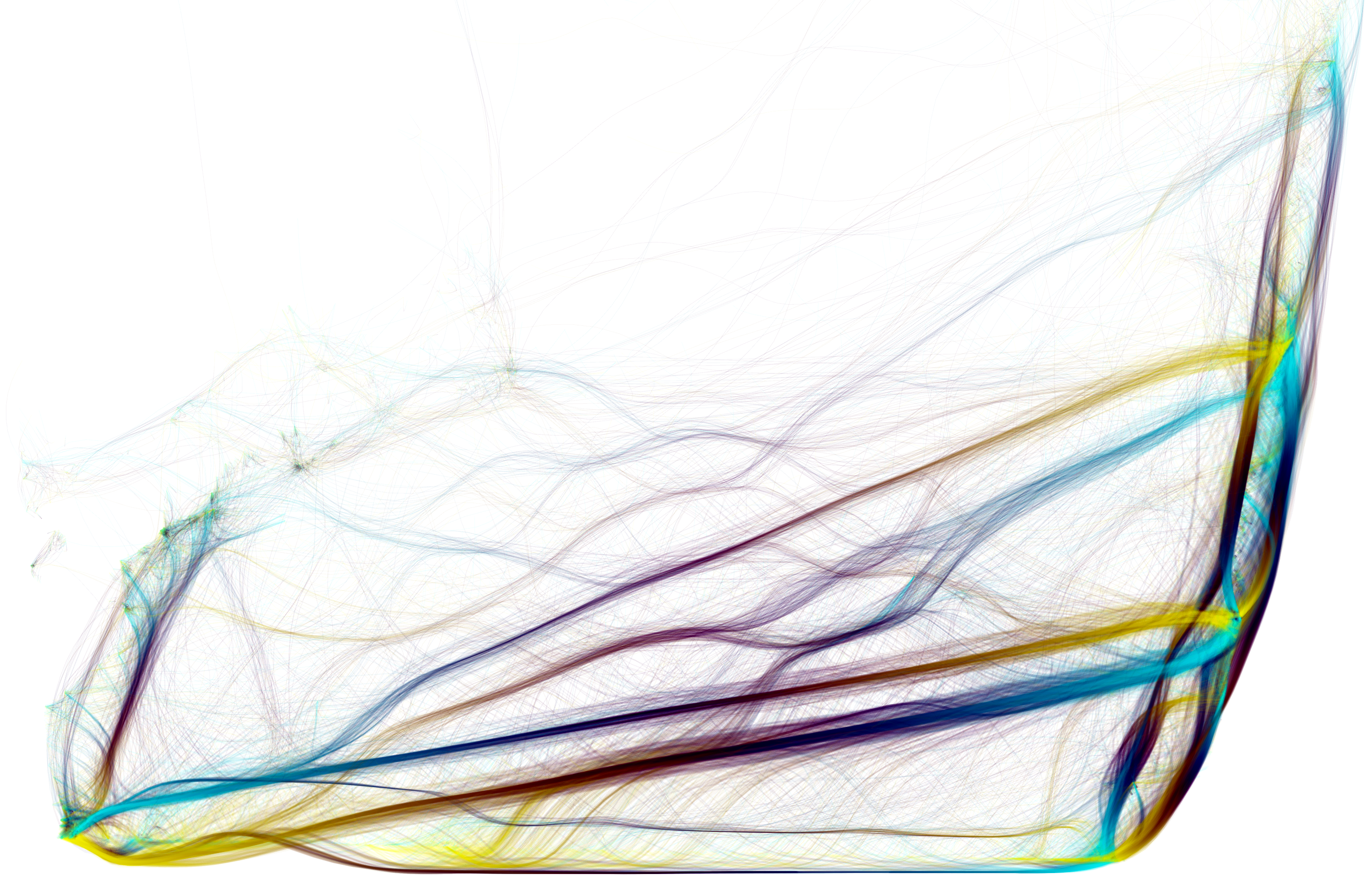}
    \caption{A kernel-based Edge Bundling visualization technique~\cite{Moura2015:3DHEB} applied over the dataset's first and last message of each unique trajectory.
    The colors are arbitrarily used to contrast the flows and ease the visualization, while the thickness of the edges represents the flow intensity.}
    \label{fig:vessels-network}
\end{figure}

\begin{figure}[b!]
    \centering
    \includegraphics[width=\linewidth]{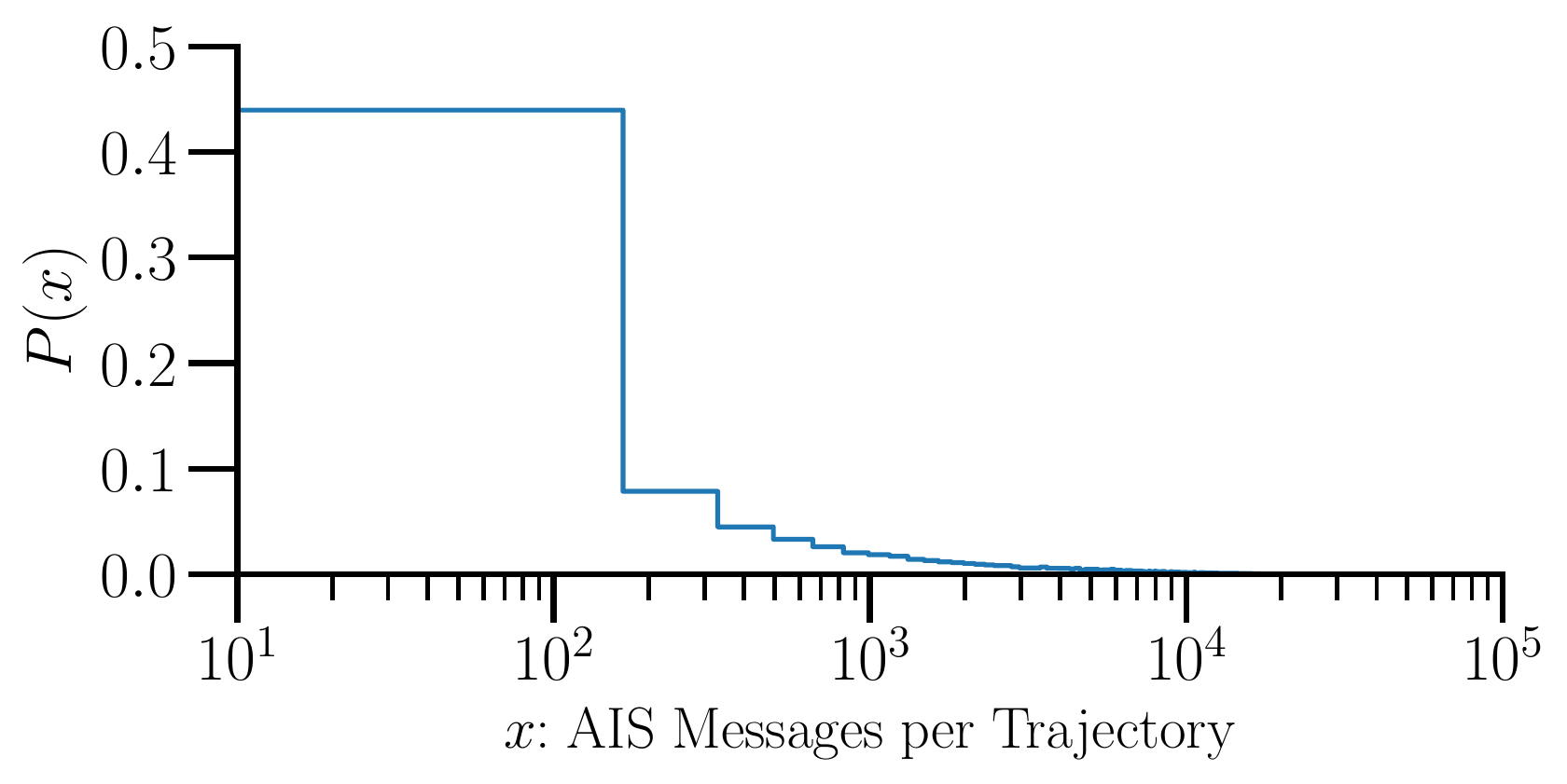}
    \caption{Probability distribution of Automatic Identification System (AIS) messages per trajectory ({\it i.e.}, vessel) in the dataset.
    It shows that most vessels have few records, and a few vessels concentrate most of the records within the dataset, a behavior comparable to a long-tail ({\it i.e.}, Pareto) data distribution.}
    \label{fig:ais-probability}
\end{figure}

The dataset used in this article comprises a portion of the Atlantic Ocean from Iceland to the south of the United States and the west of Europe to the north of Africa (see Figure~\ref{fig:ais-map}).
It consists of a private dataset provided by {\it Spire}\footnote{~\url{https://www.spire.com/}} (former {\it exactEarth}) that contains raw AIS messages of over $20,000$ vessels of different types ({\it e.g.}, cargo, tanker, fishing, and other vessels) collected from March to July 2020, resulting in about $60,000,000$ AIS messages.
It is worth noting that the vessels navigate independently and are not limited to navigating inside the bounding box containing the dataset.
In this sense, Figure~\ref{fig:vessels-network} simultaneously pictures each unique vessel's first and last appearance by an Edge Bundling visualization technique~\cite{Moura2015:3DHEB}.
Although colors are used to contrast vessels' flow, the trajectories' thickness is proportional to the recurrence of the route, indicating the intensity of the marine flow in the studied region.
These trajectories have different lengths as well as starting and ending locations, and they contain noise in the form of inaccurate information within AIS messages.
The analysis of the inaccuracy behind the AIS messages in this dataset is beyond this work's scope.

\begin{figure}[t!]
    \centering
    \includegraphics[width=\linewidth]{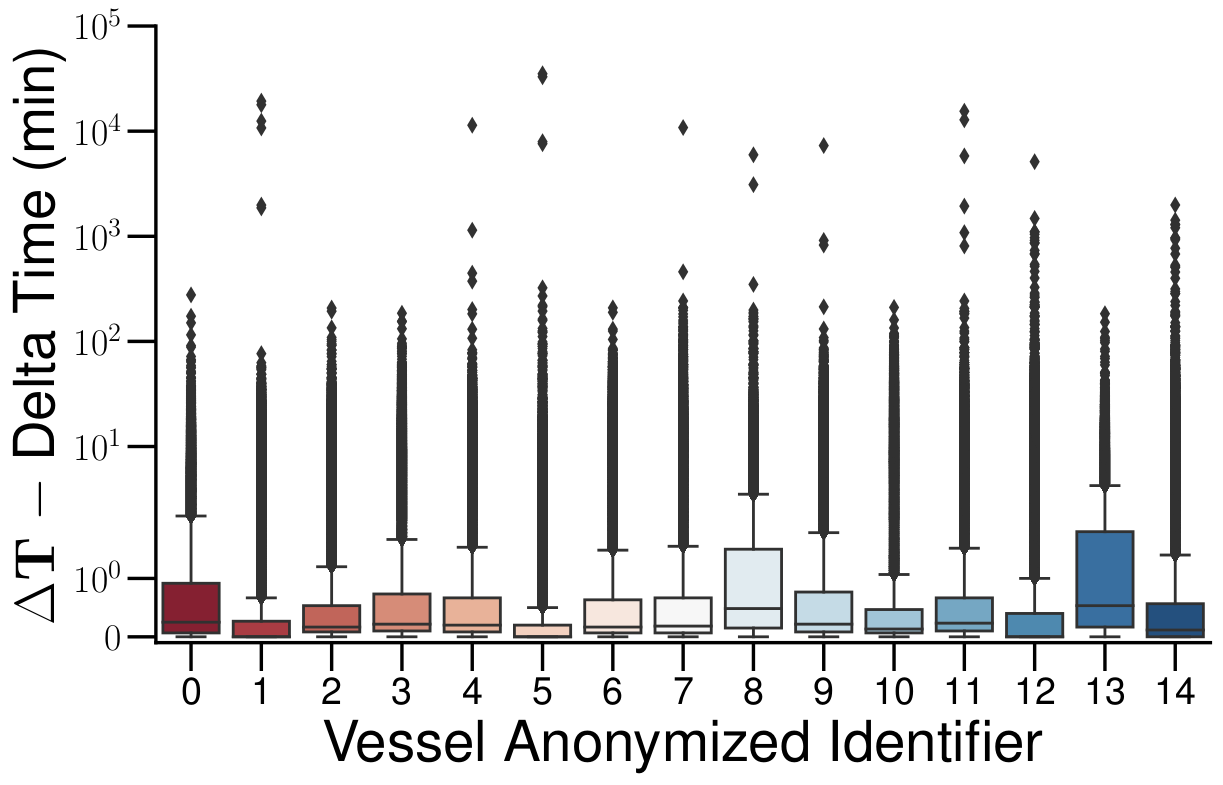}
    \caption{Interquartile Range Analysis -- IQR of delta time for fifteen different vessels ordered from the one with the most AIS messages to the one with the least.
    The analysis reveals that all vessels present a severe presence of outliers.
    An outlier indicates an irregularity related to the time elapsed between two consecutive messages, varying from a couple of seconds to a few months.}
    \label{fig:ais-time}
\end{figure}

Figure~\ref{fig:ais-probability} illustrates the probability distribution of AIS messages per trajectory.
It shows the shape of a long-tail ({\it i.e.}, Pareto) distribution, meaning that the dataset has most of its AIS messages concentrated on a small number of trajectories, and a few vessels dominate the trajectory dataset.
An unbalanced dataset such as this has a trade-off between performance and generalization.
Due to that, different data modeling approaches are required to reduce the bias of the heavily populated trajectories.
The dataset has another conspicuous feature among the trajectories, which is the irregular timing between consecutive transmitted AIS messages.
For example, Figure~\ref{fig:ais-time} illustrates the phenomenon in the form of outliers observed between consecutive messages.
The image provides the Interquartile Range Analysis (IQR) for fifteen randomly selected vessels, in which it is possible to note the extreme variance between consecutive transmissions.
Most messages are received within seconds or minutes, but there are recurrent cases where, due to transmission delays, it spiked up to a few days and even a couple of months.

% \FloatBarrier%
\subsection{Window Sampling and Scaling}
\label{sect:windowing-sampling}

AIS data are notoriously known for their long historical sequences.
Although its volume is considered an asset in many applications, its overabundance can also be detrimental, particularly in unbalanced trajectories (see Figure~\ref{fig:ais-probability}).
To increase the model's mobility pattern variability and geospatial coverage while it decreases the training time, we had to design a training technique based on temporal sampling.
However, regular AIS message sampling affects the trajectory data distribution similarly to using trajectory interpolation based on virtual AIS messages (see Section~\ref{sect:introduction}), altering the behavior underneath the transmission system that we seek to model.

\begin{figure}[b!]
    \centering
    \includegraphics[width=\linewidth]{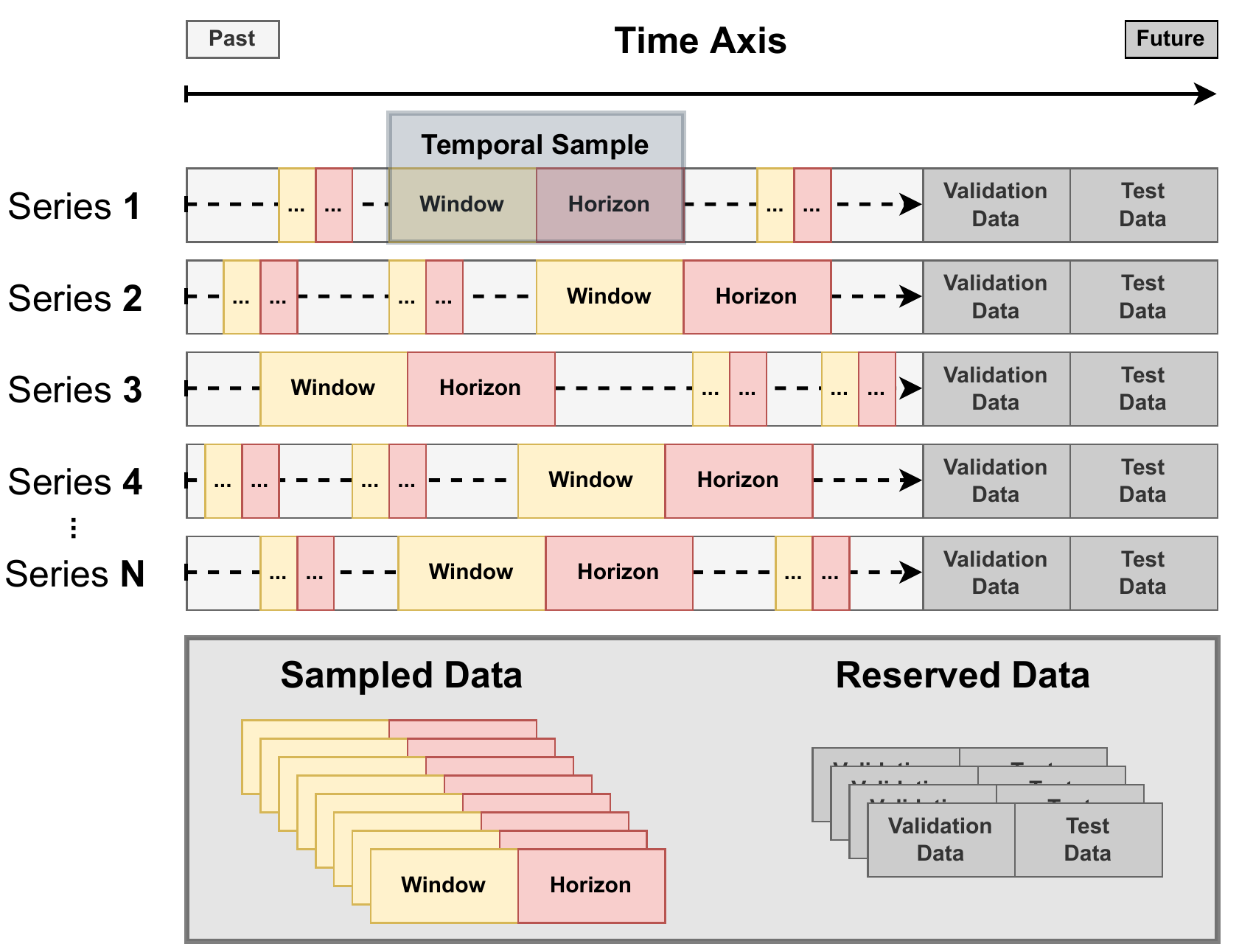}
    \caption{Temporal sampling technique designed to increase the variability of trajectories seen by the models.
    It decreases the computational training time on the entire trajectories without increasing the timing irregularity within the windows, proving segments of different trajectories and timespans.}
    \label{fig:time-sampling}
\end{figure}

To preserve the data distribution, expand the model's capacity, and still reduce training time, we transformed each trajectory into predefined temporal segments known as windows, and then we sampled the temporal windows instead of the messages contained in them (see Figure~\ref{fig:time-sampling}).
This approach preserves the course of time within the windowed AIS messages without increasing temporal irregularities inherent in message sampling.
The idea is to sample sequences from all vessels indistinctly and feed them randomly to the learning model such that the model sees segments of trajectories from multiple vessels at varying timespans and/or locations.

The data among the different sampled sequences are standardized using the $z$-score normalization, which enforces a zero mean and unit variance for all the records.
Next, the standardized samples undergo a min-max normalization to set all values on a zero-one scale.
All data transformation is applied on the variable axis shared among all the windowed samples of the dataset.
The parameters for each transformation are computed from the training set samples only and then applied to the multiple samples of the testing data.
Due to transforming the entire dataset, the models' outputs will follow an ideally similar scale.
Therefore, the output must be inversely transformed before assessing the scoring metrics.

We have set $25$ windows as the default value for the window-trajectory sampling for the experiments.
We refrain from sampling a higher number of windows because the higher the number of samples, the longer the training sessions will be.
Notice that the size of the input and output sequences scales cubically due to working on a multi-task and multivariate forecasting problem, meaning that minor variations in the number of sampling windows have the potential to quickly increase the dataset size to a point where hardware limitations will not allow moving forward with the training.
Nevertheless, aiming to increase the sampling variability, the experiments are repeated five times using different random seeds, presenting as the results the average of the experiments followed by their standard deviation.
This approach allows us to work with a high number of vessels and preserves irregular timing while increasing the reliability of the experimentation.

For training the model based on time-windowed data, the sliding window technique is a straightforward approach commonly used with sequence-and series-like data~\cite{Keogh2004:SegmentingServey}.
It works by setting a fixed-size window that slides over the temporal axis of the dataset, predicting a pre-specified number of future steps, referred to as the horizon.
Moreover, the fixed window size is known for being a highly sensitive hyperparameter~\cite{Frank2000:WindowSize, Frank2001:SlidingWindow}, which leads us to set it before the experiments by considering the domain of the data and the dataset itself~\cite{Spadon2021:ReGENN}, besides hardware limitations that come with working on large datasets made of long sequences from multiple vessel trajectories.
The window $w$ and horizon $s$ sizes used for experimentation along with the paper are presented as follows in three complexity categories:

\begin{itemize}
    \item $w=15~\&~s=05$ --- \textit{\textbf{low complexity}};
    \item $w=15~\&~s=25$ --- \textit{\textbf{medium complexity}}; and,
    \item $w=30~\&~s=50$ --- \textit{\textbf{high complexity}}.
\end{itemize}

For two out of three categories, the window sizes were set to be smaller than the horizon to increase the difficulty of the forecasting task, which will look to fewer past events for forecasting a larger horizon.
However, forecasting sequences larger than the ones we used might increase the uncertainty of the forecasting process by stacking the error of the sequentially forecasted messages, possibly generating an output that no longer represents the target network.
For example, assuming a dataset has $1,000$ different trajectories, the window size is $30$, and the horizon size is $50$.
The model will digest $(30 \times 25) \times 1,000$ AIS messages in a single iteration over the entire dataset and provide as output $(50 \times 25) \times 1,000$ AIS messages.
Knowing that our dataset has around $20,000$ different trajectories (see Section~\ref{sect:spatial-coverage}), our input/output has a $20$ times larger magnitude.
Therefore, in the \textit{\textbf{low complexity}} experiment, throughout training and testing the model outputs $2.5M$ AIS messages, $12.5M$ in the \textit{\textbf{medium complexity}}, and $25M$ in the \textit{\textbf{high complexity}} case.
These messages are processed in mini-batches, meaning they are not processed at once but in hundred of vessels instead.

\subsection{Optimization Strategy}
\label{sect:optimization-strategy}

The proposed model is trained using a mini-batch-based optimization strategy.
In such a strategy, the algorithm iterates over the different samples of the dataset, feeding the network with mini-batches of different windowed data, repeating the process for all samples in random order.
Feeding the neural network model with randomly ordered windowed data is imperative to achieve maximum generalization.
Otherwise, the model could walk towards a {\it local optimum} due to recurrently focusing on samples of the same data distribution at the early beginning of the training.
The network parameters are shared among the dataset and optimized towards the {\it minima} of the loss function.
We used AdamW~\cite{Loshchilov2019:AdamW} as the optimizer, a gradient descent-based algorithm.
AdamW is a standard optimizer for sequence and series forecasting tasks, a variant of Adam~\cite{Kingma2015:Adam} with improved decoupled weight regularization.
As the optimization criterion, we used the Hyperbolic Tangent Error (HTE), which is defined as:
{
    \begin{equation}
        \minimize_{\myvec{\Omega}} \frac{1}{N} \sum_{i=1}^{N} \left(y_i - \hat{y_i}\right) \times tanh\left(y_i - \hat{y_i}\right)\text{,}
    \end{equation}
}

\noindent%
where $\myvec{\Omega}$ are the network parameters, $N$ is the number of mini-batches, $y$ is the ground truth, and $\hat{y}$ is the prediction.

The HTE behaves similarly to the traditional Mean Absolute Error (MAE), and both are less sensitive to outliers, but HTE allows for more refined results generalization in the face of the problem constraints observed in the trajectory network.
The significant difference between them is that the derivative used to compute the gradients and update the weights is a step function for the MAE and a non-linear function for the HTE.
The optimization criterion is calculated from the full content of the AIS messages and not only the trajectory itself.
In such a way, the overall error is a compound function of the individual errors of each variable in the massage, which are all on the same scale (see Section~\ref{sect:windowing-sampling}).
We aim to find a model that near-optimally minimizes the error of simultaneously forecasting the continuous variables in AIS messages of multiple trajectories of different vessel types.

Due to working with noisy and non-prepossessed AIS data, we inserted a clipping function that enforces the boundaries of the AIS message information they represent ({\it i.e.}, longitude - $\rho$, latitude - $\omega$, $\Delta\mathrm{T}$, COG - $\epsilon$, and SOG - $\mu$) after the model output computation and before computing the loss function.
The clipping function first undoes the min-max and $z$-score normalization and then enforces the following constraints:
\begin{itemize}
    \item[] ~~~~$\rho = min(max(-180, \rho), 180) \equiv \rho \in \left[-180, 180\right]$
    \item[] ~~~~$\omega = min(max(-90, \omega), 90) \equiv \omega \in \left[-90, 90\right]$
    \item[] $\Delta\mathrm{T} = min(max(0, \Delta\mathrm{T}), \infty) \equiv \Delta\mathrm{T} \in \left[0, \infty\right[$
    \item[] ~~~~$\epsilon = min(max(0, \epsilon), 360) \equiv \epsilon \in \left[0, 360\right]$
    \item[] ~~~~$\mu = min(max(0, \mu), \infty) \equiv \mu \in \left[0, \infty\right[$
\end{itemize}
\noindent%
We have used the same clipping function on the entire dataset before computing the evaluation metrics for off-the-shelf algorithms not trained using our network training pipeline.

\subsection{Evaluation Metrics}
\label{sect:evaluation-merics}

{
In addition to the network optimization criterion, the results are presented with the aid of the Relative Percentage Difference (RPD) and Root Mean Squared Error (RMSE):

\begin{equation}
    \text{RPD} = \frac{2}{N}~\sum^{N}_{i=1} \frac{y_i - \hat{y}_i}{|y_i| + |\hat{y}_i|}
\end{equation}

\begin{gather}
    \text{RMSE} = \sqrt{\frac{1}{N}~\sum^{N}_{i=1}\left(y_i - \hat{y}_i\right)^2}
\end{gather}

\noindent%
where $y$ is the ground truth, $\hat{y}$ is the model prediction, and $N$ is the number of mini-batches.
The RMSE, based on the square root, is used to evaluate the model in the face of larger values, which in the case of the vessel trajectory network dataset are known to be outliers.
Alternatively, RPD is a signed expression that compares the difference between the values and their average magnitude.
The Hyperbolic Tangent (HTE), used as the loss function, and the RMSE are bound to $[0, \infty[$, where $0$ indicates a perfect model, and greater values indicate otherwise.
It is noteworthy that, along with the results, predicting outliers is not the model's objective; a robust model will show satisfactory generalization among the median values given by the HTE regardless of the outliers noted by the RMSE.
The RPD is bounded to $[-2, 2]$, where the more accurate the model is, the closer to zero it will be. In this sense, negative values mean the predictions are generally lower in value than the ground truth, and positive values indicate they are generally greater.
Accordingly, we seek a model that achieves an average as close to zero and a low standard deviation as possible among HTE and RPD but not necessarily low RMSE values.
}

\subsection{Network Architecture}
\label{sect:network-architecture}

The neural network proposed for modeling the vessel trajectory network under irregular timing constraints and in the face of different data distributions consists of two sequential single-directed and single-layered long-short-term memory ({\it i.e.}, LSTM~\cite{Hochreiter1997:LSTM}) cells that operate with the aid of a one-dimensional convolution ({\it i.e.}, Conv1D~\cite{Abdeljaber2017:Conv1D}) feature-extraction layer before each LSTM, while simultaneously leveraging a linear feed-forward shortcut connecting the network input to the output in a residual-like connection~\cite{tai2017image} with trainable parameters.
Each triplet of convolution, recurrent encoding, and sequential decoding is referred to as a block, having independent weights but being trained together, whereby the first is labeled as $\alpha$ and the second as $\omega$.

In such a case, after the windowing and window-sampling preparation processes (see Section~\ref{sect:windowing-sampling}), the data from the multiple trajectories is fed to a convolutional layer.
In this layer, the multiple features existing within the windowed trajectories in a mini-batch ({\it i.e.}, input planes) will be combined into an intermediate tensor representation containing the hidden features that arise from the cross-correlation between the weights and the input planes.
As a result, the hidden features will have the temporal axis dilated (or contracted) to match the number of output channels of the convolutional layer, initially set to be the window size $w$.
{
Due to leveraging a single-dimension convolution, the variables will be convolved only with themselves and never with the other variables within the message. This means that a contracted sequence of messages is a smaller representation of the trajectory, similarly to the output of a series reduction algorithm. On the other hand, having an expanded output of the input sequence can be understood as an interpolated segment of the input trajectory. These messages, however, arise from the hidden weights of the network and have no straightforward meaning as the original messages; therefore, we refrain from further comparing the original trajectory with the one arising from the hidden weights of the proposed neural network.
}

The one-dimensional convolution can be defined as:

\begin{equation}
    x_t^{\alpha} = \left(\sum_{p = 0}^{\mathcal{I} - 1} \mymtx{W}^{(p)}_{\mathcal{O}} \star x^{(p)}\right) + \myvec{b}_{\mathcal{O}} \text{,}
    \label{eq:CNN}
\end{equation}

\noindent%
where $\mymtx{W} \in \mathbb{R}^{\mathcal{O} \times \mathcal{I} \times k}$ is the weights, $\myvec{b} \in \mathbb{R}^{\mathcal{O}}$ the bias, $\star$ the cross-correlation operator, $t$ the time instant indicator, $k$ is the kernel size, $\mathcal{O}$ the number of output channels, and $\mathcal{I}$ the number of input channels --- bounded to a sequence of size $w$, the sliding window's size.
The output of the convolutional layer will be the hidden features with a temporal dimension matching the number of output channels.

Next, the hidden features extracted by the convolutional layer go through the first LSTM of the network, defined as:
{%
    \begin{align*}
        i_t^{\alpha} &= \sigma\left(\left(\mymtx{W}_{ii} \cdot x_t^{\alpha} + \myvec{b}_{ii}\right) + \left(\mymtx{W}_{hi} \cdot h_{t-1} + \myvec{b}_{hi}\right)\right) \\
        f_t^{\alpha} &= \sigma\left(\left((\mymtx{W}_{if} \cdot x_t^{\alpha} + \myvec{b}_{if}\right) + \left(\mymtx{W}_{hf} \cdot h_{t-1} + \myvec{b}_{hf}\right)\right) \\
        g_t^{\alpha} &= tanh\left(\left(\mymtx{W}_{ig} \cdot x_t^{\alpha} + \myvec{b}_{ig}\right) + \left(\mymtx{W}_{hg} \cdot h_{t-1} + \myvec{b}_{hg}\right)\right) \\
        o_t^{\alpha} &= \sigma\left(\left(\mymtx{W}_{io} \cdot x_t^{\alpha} + \myvec{b}_{io}\right) + \left(\mymtx{W}_{ho} \cdot h_{t-1} + \myvec{b}_{ho}\right)\right) \\
        c_t^{\alpha} &= \left(f_t~\circ~c_{t-1}\right) + \left(i_t~\circ~g_t\right) \\
        h_t^{\alpha} &= o_t~\circ~tanh\left(c_t\right) \tag{\stepcounter{equation}\theequation} \text{,}\\
    \end{align*}
    % \raisetag{15pt}
    \label{eq:LSTM-1}
}

\vspace{-.8cm}\noindent%
where $\mymtx{W}_{i}, \mymtx{W}_h \in \mathbb{R}^{\mathcal{O} \times \mathcal{O}}$ are the weights and $\myvec{b} \in \mathbb{R}^{\mathcal{O}}$ the bias to be learned, $i_t^{\alpha}$ is the input and update gate's activation vector, $f_t^{\alpha}$ the forget gate's activation vector, $g_t^{\alpha}$ the cell gate, $o_t^{\alpha}$ the output gate's activation vector, $c_t^{\alpha}$ the cell state vector, $h_t^{\alpha}$ the hidden state vector, and $\sigma$ the sigmoid activation function, $\circ$ the Hadamard product.
Next, the last hidden state vector of the first LSTM cell, {\it i.e.}, $h_t^{\alpha}$, is then fed to a non-linear feed-forward decoder that will convert the hidden-size dimension of the data into the expected output size regarding only the temporal dimension formalized as follows:
{%
\begin{align*}
    \Tilde{x}_t^{\alpha} &= \mathtt{ReLU}\left(\delta\left(\mymtx{W}_m \cdot h_m^{\alpha} + \myvec{b}_m\right)\right)\tag{\stepcounter{equation}\theequation}
    % \label{eq:decoder}
\end{align*}
}%
where $\mymtx{W}_m \in \mathbb{R}^{\mathcal{O} \times m}$ is the weights, $\myvec{b}_m \in \mathbb{R}^{m}$ the bias, $m$ is the number of variables, and $\delta$ the dropout operation.

The previous network layer's block will use the set of gates and memory of the LSTM cell to unfold the sequences in the hidden features created from the cross-correlation operation.
It will incorporate traces of the multiple data distributions in the internal weights yielding an intermediate result.
Due to the increased complexity of working on a multi-task multivariate forecasting task, a single network block showed not to be not enough.
Therefore, we permuted the tensor exposing the variable axis to a different block for re-coding the temporal axis while learning intricacies from the variables instead.
Using this approach, the first block learns how the variables of the AIS sequence change through time, while the second learns how time changes through the intermediate hidden weights representing the variables.
As a result, the output of the previous block, {\it i.e.}, $\Tilde{x}_t^{\alpha}$, is then in-sequence stacked to a second block formalized as follows:

\paragraph*{Conv1D}

\begin{align*}
    x_t^{\omega} &= \left(\sum_{p = 0}^{\mathcal{I} - 1} \mymtx{W}^{(p)}_{\mathcal{O}} \star x^{(p)}_{\alpha}\right) + \myvec{b}_{\mathcal{O}} \text{,} \tag{\stepcounter{equation}\theequation}
\end{align*}

\paragraph*{LSTM Encoder}

\begin{align*}
    i_t^{\omega} &= \sigma\left(\left(\mymtx{W}_{ii} \cdot x_t^{\omega} + \myvec{b}_{ii}\right) + \left(\mymtx{W}_{hi} \cdot h_{t-1} + \myvec{b}_{hi}\right)\right) \\
    f_t^{\omega} &= \sigma\left(\left((\mymtx{W}_{if} \cdot x_t^{\omega} + \myvec{b}_{if}\right) + \left(\mymtx{W}_{hf} \cdot h_{t-1} + \myvec{b}_{hf}\right)\right) \\
    g_t^{\omega} &= tanh\left(\left(\mymtx{W}_{ig} \cdot x_t^{\omega} + \myvec{b}_{ig}\right) + \left(\mymtx{W}_{hg} \cdot h_{t-1} + \myvec{b}_{hg}\right)\right) \\
    o_t^{\omega} &= \sigma\left(\left(\mymtx{W}_{io} \cdot x_t^{\omega} + \myvec{b}_{io}\right) + \left(\mymtx{W}_{ho} \cdot h_{t-1} + \myvec{b}_{ho}\right)\right) \\
    c_t^{\omega} &= \left(f_t~\circ~c_{t-1}\right) + \left(i_t~\circ~g_t\right) \\
    h_t^{\omega} &= o_t~\circ~tanh\left(c_t\right) \tag{\stepcounter{equation}\theequation} \text{,}
\end{align*}

\paragraph*{Linear Decoder}

\begin{align*}
    \hat{y}^{\omega} = \mymtx{W}_n \cdot h_t^{\omega} + \myvec{b}_n \tag{\stepcounter{equation}\theequation}
\end{align*}

\noindent
where the weights and bias for the \texttt{Conv1D} and the \texttt{LSTM Encoder} follow the exact dimensions as the first block but not the last linear layer where $\mymtx{W}_n \in \mathbb{R}^{\mathcal{O} \times n}$ is the weights, $\myvec{b}_n \in \mathbb{R}^{n}$ the bias, $n$ is the number of variables.
There is no dropout nor activation function applied to this block's output.

As previously mentioned, due to the neural network consistently losing the scale of the output compared to the dataset's input, we leveraged an additional \texttt{Linear} layer that works in parallel with the rest of the architecture.
Such a linear layer is comparable to an \textit{Autoregressive} component~\cite{Lai2018:LSTNet}, in which no non-linearity is applied to either the input or output of the layer.
The component works by restoring the scale of the data that, due to subsequent operations and non-linearities, makes the output tend to zero.
The following gives the final output of the proposed neural network model:

\begin{align*}
    \hat{y} = \left(\mymtx{W}_{ar} \cdot x + \myvec{b}_{ar} \tag{\stepcounter{equation}\theequation}\right) + \hat{y}^{\omega}
\end{align*}

\subsubsection*{Baselines}

We considered over $60$ different traditional and state-of-the-art algorithms as a baseline.
This experimental set includes machine and deep learning models adapted for the trajectory AIS transmission task, using the training preparation steps described in Sections~\ref{sect:spatial-coverage} and~\ref{sect:windowing-sampling}.

The machine learning algorithms (see supplemental material for a complete list) come from open-source libraries, {\it e.g.}, scikit-learn~\cite{Pedregosa2011:Scikit}, scikit-multiflow~\cite{Montiel2018:Skmultiflow}, scikit-extra\footnote{~Available at~https://bit.ly/3tqPg3f.}, lightning~\cite{Blondel2016:Lightning}, and polylearn\footnote{~Available at~https://bit.ly/3KfGVFw.}.
Other estimators, such as CatBoost~\cite{Liudmila2018:CatBoost}, XGBoost~\cite{Chen2016:XGBoost}, and LGBM~\cite{Ke2017:LGBM}, have their dedicated open-source implementation, which was preferred over others.
Notably, most of these out-of-shelf algorithms operate on a single-or multi-output sample space.
However, even the more adaptable algorithm lacks straightforward support for multi-output and multi-task forecasting problems.%, which are the case of multiple vessels' trajectories describing many time-varying variables.

Therefore, we adapt the single-output algorithms into multi-output ones using a {\it Regression Chain} mechanism\footnote{~Available at~https://bit.ly/3hBfxTA.}.
This technique combines multiple single-output estimators of the same algorithm in the order specified by the chain, having one different estimator for each inferred horizon unit, in which the previous estimator feeds the following estimator~\cite{Melki2017:Chains}.
However, even in a chained pipeline, these estimators cannot simultaneously focus on the multiple samples and variables.
Therefore, the problem was split into smaller parts, allowing the chained single-output and multi-output algorithms to focus on a single variable shared among all trajectories simultaneously, repeating the process for each variable in the dataset and then averaging the final results.

This approach simplifies the inference process, as the algorithms are now centered on a single variable per time instead of being required to forecast all of them simultaneously.
However, it is essential to note that, although the problem is more straightforward in terms of the number of variables simultaneously predicted, there is less interaction between multivariate samples, which might mean these estimators learn a limited amount of inter-variable features when compared to multi-output and multi-task ones.

In order to ease the understanding of the inference limitation of the baseline algorithms, along with the experiments, we have symbol-encoded them using the subsequent scale:

\begin{itemize}
    \item[\faBullseye] Represents single-output algorithms;
    \item[\faDotCircleO] Indicates multi-output algorithms; and,
    \item[\faCircleO] Consists of multi-output and multi-task algorithms.
\end{itemize}

Specifically, among the deep learning baselines, we have used a different set of network architectures adapted and re-implemented for specifically handling the data from the vessel trajectory network.
Related to Recurrent Neural Networks, we have conducted experiments with Elman's RNN \cite{Elman1990:RNN}, GRU~\cite{Chung2014:GRU}, and LSTM~\cite{Hochreiter1997:LSTM}.
For Auto-Encoders, we have simplified ReGENN~\cite{Spadon2021:ReGENN} for a bi-dimensional input, in which the Transformer Encoder~\cite{Vaswani2017:Attention} is used to extract an encoded representation from the input features, and an LSTM is used to decode such a representation into the horizon.
Regarding Convolutional Neural Networks~\cite{Lecun1998:CNN}, we experimented on a temporal CNN with a single-dimension convolutional layer followed by a feed-forward layer that translates the output channels resulting from the cross-correlation operation into the horizon.
We experimented with a feed-forward network for accessing the results on a linear multi-output and multi-task estimator and included an additional set of deep learning baselines, which are the highway networks~\cite{Srivastava2015:Highway, Zilly2017:RecurrentHighway}.
Note that these estimators might lose the significance of the output scale predictions compared to the input when the information is propagated throughout the network repeatedly.

\subsubsection*{Hyperparameter Tuning}
Along with the experimentation, we use the default hyperparameters for all algorithms.
More specifically, for the machine-learning baselines, the hyperparameters come from the open-source library where they are included (see the supplementary material for details), and for the deep-learning ones, PyTorch's defaults unless specified.
We used a gradient norm-clipping of $1.0$, a learning rate of $1e^{-3}$, $10\%$-probability dropout, and a learning rate scheduler to reduce the learning rate by a fifth every three stalled epochs.
For the CNNs, specifically, we have used a fixed kernel size of $3$, padding the input with a stride of $1$, so the output has the same shape as the input but with an increased number of output channels ({\it i.e.}, $128$) when compared to the input channel, which matches the size of the window.
For the recurrent networks, including our model, we set a pre-fixed hidden size of $128$ for all the experiments.

As part of the results, we show how our network behaves when we change the number of output channels of the convolutional layer (between $8$, $16$, $32$, $64$, and $128$ channels) and also when we vary the recurrent layer (between Elman's RNN, GRU, and LSTM) in addition to their number of stacked layers ranging between $1$ and $3$.
All the experiments were repeated five times with different random seeds ({\it i.e.}, $2021$, $2121$, $2221$, $2321$, and $2421$) to increase the variability of the sampled data during the experimentation and the order that the networks will see the samples (see Section~\ref{sect:windowing-sampling}).

\subsubsection*{Computer Environment}
The experiments related to machine-learning algorithms were conducted on a Linux-based system with $80$ CPUs and $504$ GB of RAM.
The ones related to deep learning were carried out on another Linux-based system with $48$ CPUs, $126$ GB of RAM, and a GeForce A100 $40$ GB (Ampere).

\subsubsection*{Reproducibility}
The dataset used in this paper is not available to the general public for download due to being a private dataset owned by {\it Spire}.
However, aiming at the reproducibility of the results, we provide the source code, the snapshot of the proposed network on {\it GitHub}\footnote{~Available at~https://github.com/gabrielspadon/ais-transmissions.}, guiding the user on how the inference process should be carried out on a sample dataset.

\section{Results \& Discussion}
\label{sect:results-discussion}

\subsection*{Performance over complexity scenario}
This section describes the results considering the different experimental complexity setups as highlighted by Section~\ref{sect:windowing-sampling}.
Due to the gradual transition in the problem complexity containing different window and horizon sizes, many tested algorithms presented divergent behavior.
In these cases, the algorithms could not answer all the experimental settings given their required resources and computing time inherent to the scale of our dataset (see Section~\ref{sect:spatial-coverage}).
The supplementary material presents a comprehensive list of all the algorithms while highlighting those removed from the pipeline.

\begin{figure}[!t]
    \centering
    \includegraphics[width=\linewidth]{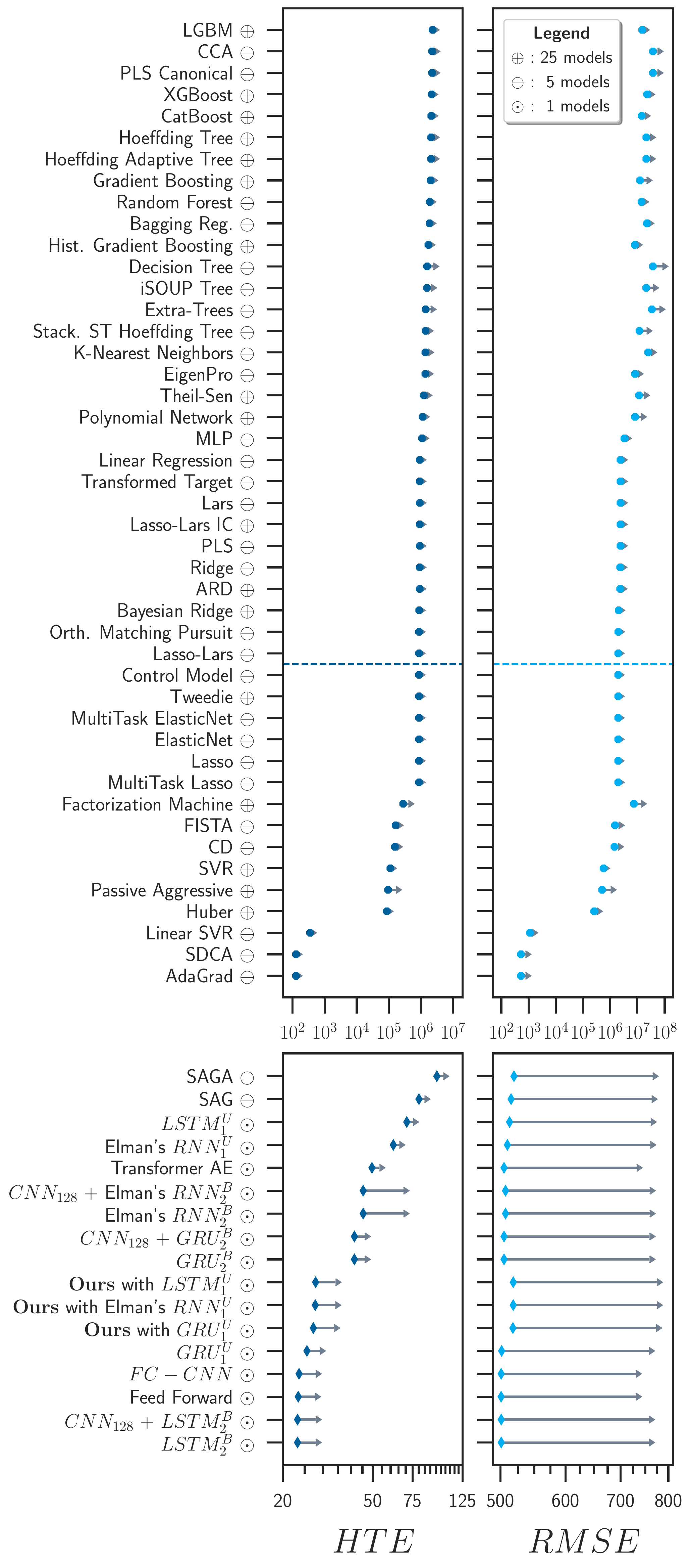}
    \caption{Performance estimation and comparison among different algorithms used for modeling the vessel trajectory network considering the low complexity case where algorithms look for the last $15$ messages to predict the subsequent $5$ messages.
    The performance assessment is based on the Hyperbolic Tangent Error (HTE) and the Root Mean Squared Error (RMSE).
    The experiments were conducted with algorithms on their out-of-the-box version with no hyperparameter optimization.
    Specifically, among the neural networks, we use $U$ as a superscript to indicate a single-directed model, $B$ for double-directed, and the subscript numbers as the number of stacked recurrent cells.}
    \label{fig:result-11}
\end{figure}

\begin{figure}[!b]
    \centering
    \includegraphics[width=\linewidth]{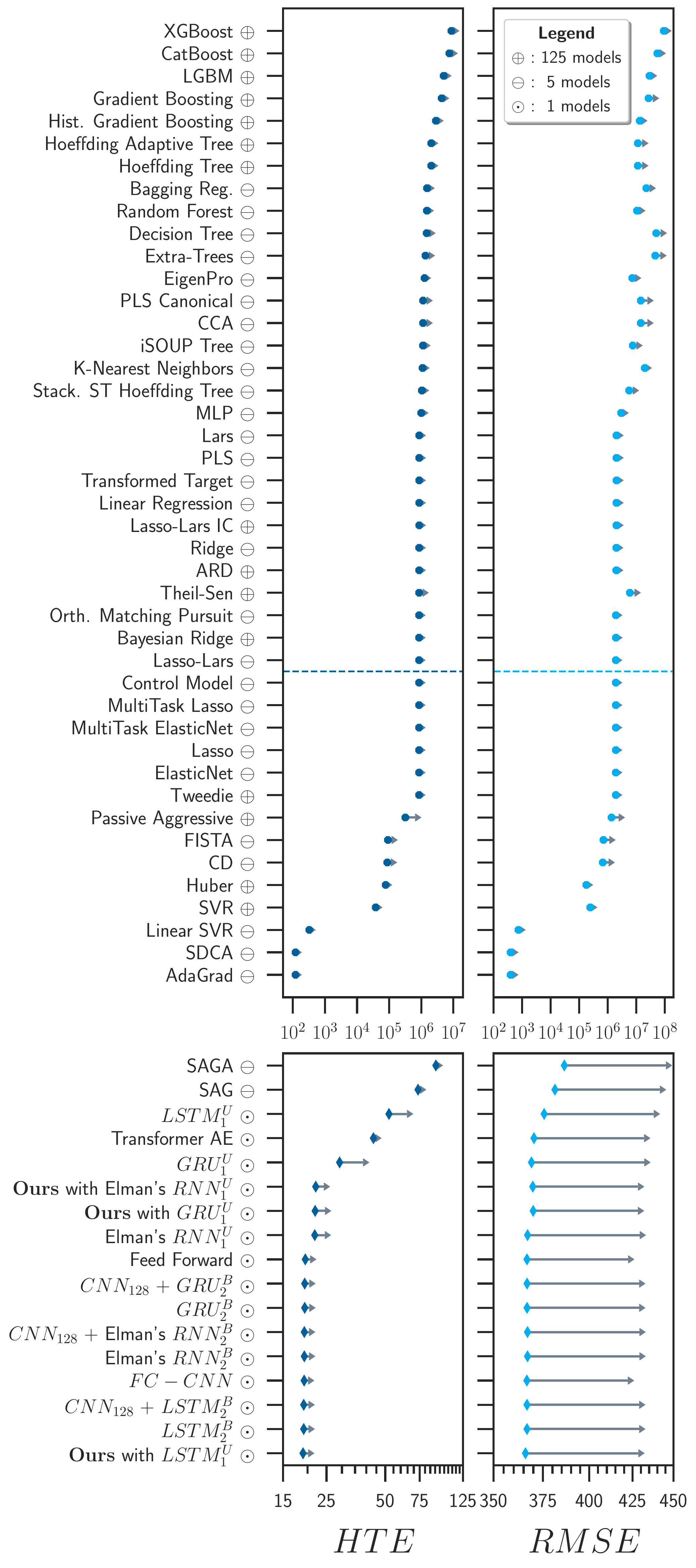}
    \caption{Performance estimation and comparison among different algorithms used for modeling the vessel trajectory network considering the medium complexity case where algorithms look for the last $15$ messages to predict the subsequent $25$ messages.
    The performance assessment is based on the Hyperbolic Tangent Error (HTE) and the Root Mean Squared Error (RMSE).
    The experiments were conducted with algorithms on their out-of-the-box version.
    Specifically, among the neural networks, we use $U$ as a superscript to indicate a single-directed model, $B$ for double-directed, and the subscript numbers as the number of stacked recurrent cells.
    The estimators used the same dataset, but the deep learning baselines leveraged our proposed model's HTE loss function and further training adaptation.}
    \label{fig:result-21}
\end{figure}

Among all the machine learning baselines, we included a {\it Control Model}, which, like the other machine learning models, will use the {\it Regression Chain} mechanism to infer over the data.
Such an inference is based on the average window size that feeds the algorithm.
Such a model divides the first set of estimators into two further pieces, as denoted by the colored dashed lines in Figures~\ref{fig:result-11}, \ref{fig:result-21}, and \ref{fig:result-31}.
This division means that estimators above the dashed line performed worse than the average, while those below performed better.
The average of the input AIS messages describes vessels nearly stalled, {\it i.e.}, in a back-and-forth moving pattern, during the horizon duration despite the other features among the AIS messages.
Performing worst than the average is a piece of evidence that they cannot represent the multiple patterns arising from different trajectories of different vessel types.

The models below the dashed lines concentrate on some high-scoring machine learning algorithms and the neural network models used for experimentation.
It is possible to notice that the neural networks are in first-placed positions, while the low-scoring among the high-scoring ones are chained out-of-the-shelf machine-learning models.
This is because neural networks can cope with the multi-task multivariate nature of our problem (see Section~\ref{sect:problem-formalization}).

\subsubsection*{Low complexity case}
The experiments start with the low complexity case, where for a fixed input of $15$ messages, we are looking to predict the subsequent $5$ messages for multiple vessels simultaneously. The low complexity of such an experiment comes from the fact that most of the AIS messages among the data, as shown in Figure~\ref{fig:ais-time}, have a low delta time between consecutive transmitted messages. Due to that, the frequency of consecutive messages is higher, which is usually related to terrestrial-based AIS messages. In this sense, for a short period, the variability in the trajectory, speed, and positioning of vessels tend to change very little if not remain nearly constant in the case of COG and SOG. In this case, simpler models, such as a Multi-Layer Perceptron (MLP), {\it i.e.}, Feed-Forward, showed more effectiveness than our solution, as well as bidirectional double-layered LSTM and its Temporal CNN version.

\begin{figure}[!b]
    \centering
    \includegraphics[width=\linewidth]{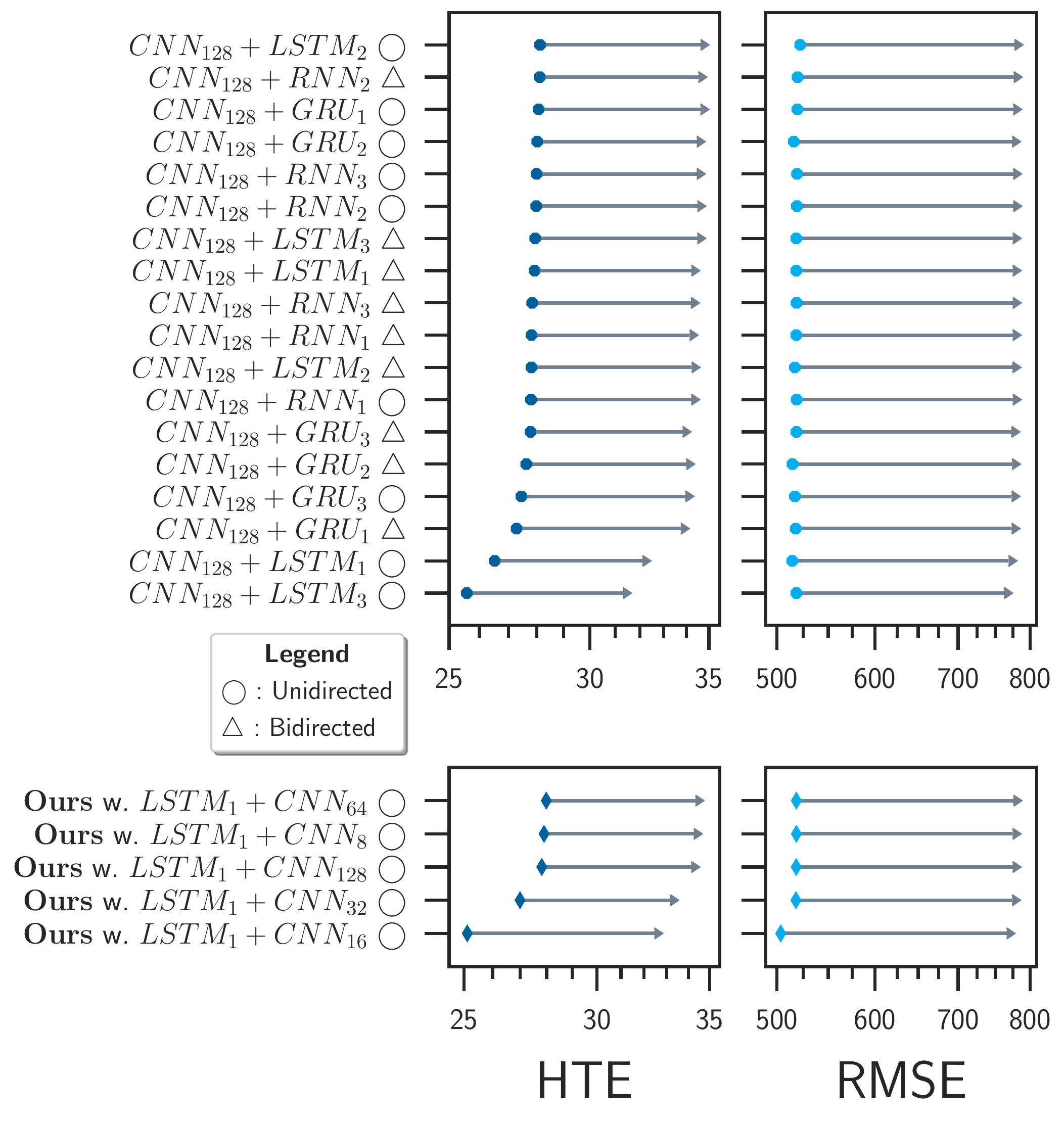}
    \caption{Impact analysis of different Recurrent Neural Networks (RNN) working in different directions and with a varying number of stacked layers compared to our proposed model for modeling the vessel trajectory network where algorithms look for the last $15$ messages to predict the subsequent $5$ messages.
    In addition to analyzing the impact of the output channels from the Convolutional Neural Network (CNN) in our proposed modeling approach.
    The performance assessment is based on the Hyperbolic Tangent Error (HTE) and the Root Mean Squared Error (RMSE).}
    \label{fig:result-12}
\end{figure}

In Figure~\ref{fig:result-12}, we further stress our model, showing how it behaves when leveraging different Recurrent Neural Networks over one or more stacked recurrent cells.
The image reveals that stacking LSTMs can increase the performance of the model, as it will be able to capture more nuanced relationships arising from the trajectories.
However, that would mean the RNN unit of the model would have up to six times more parameters than it initially had, implicating longer training sessions and potential scalability issues.
Contrarily, in the lower half of the image, we show that by using an LSTM as the RNN architecture, decreasing the hidden size and channels of the LSTMs and CNNs simultaneously on our proposed blocks can reduce the number of parameters and achieve increased performance.
In such a way, the proposed solution lies in the standard deviation of the top performers.

\subsubsection*{Medium complexity case}
Subsequently, in Figure~\ref{fig:result-21}, we analyze the medium complexity case, where we are looking to forecast the following $25$ messages using the $15$ previous messages transmitted in sequence by the vessels.
In contrast with the low complexity case, for this one, we have one further model that performed worse than the control model, the same holds for the high-complexity case.
The reason is the increased complexity of handling longer sequences and more data.
Such behavior was expected because, for small sequences, such models could not capture the interactions from the chained regression forecasting pipeline.
Further fine-tuning the hyperparameters of each estimator in the chain could undoubtedly improve the forecasting process and yield better results.
However, as the number of estimators per model ensemble increases with the model's complexity, such a modeling perspective would turn into an extensively laborious task not covered in this work.

In the lower half of Figure~\ref{fig:result-21}, MLP shows divergent behavior than previously seen because as the sequences start to get large, the more the probability of increasing the temporal gaps between consecutive AIS messages.
In such a case, the recurrency within the RNNs is better leveraged, supporting that our proposed solution achieves increased performance than other models.
This can be seen when analyzing the RMSE values. Although the variation is small, our model has a lower RMSE value, achieving slightly better results when larger temporal gaps are present in the sequence of messages.

\begin{figure}[!ht]
    \centering
    \includegraphics[width=\linewidth]{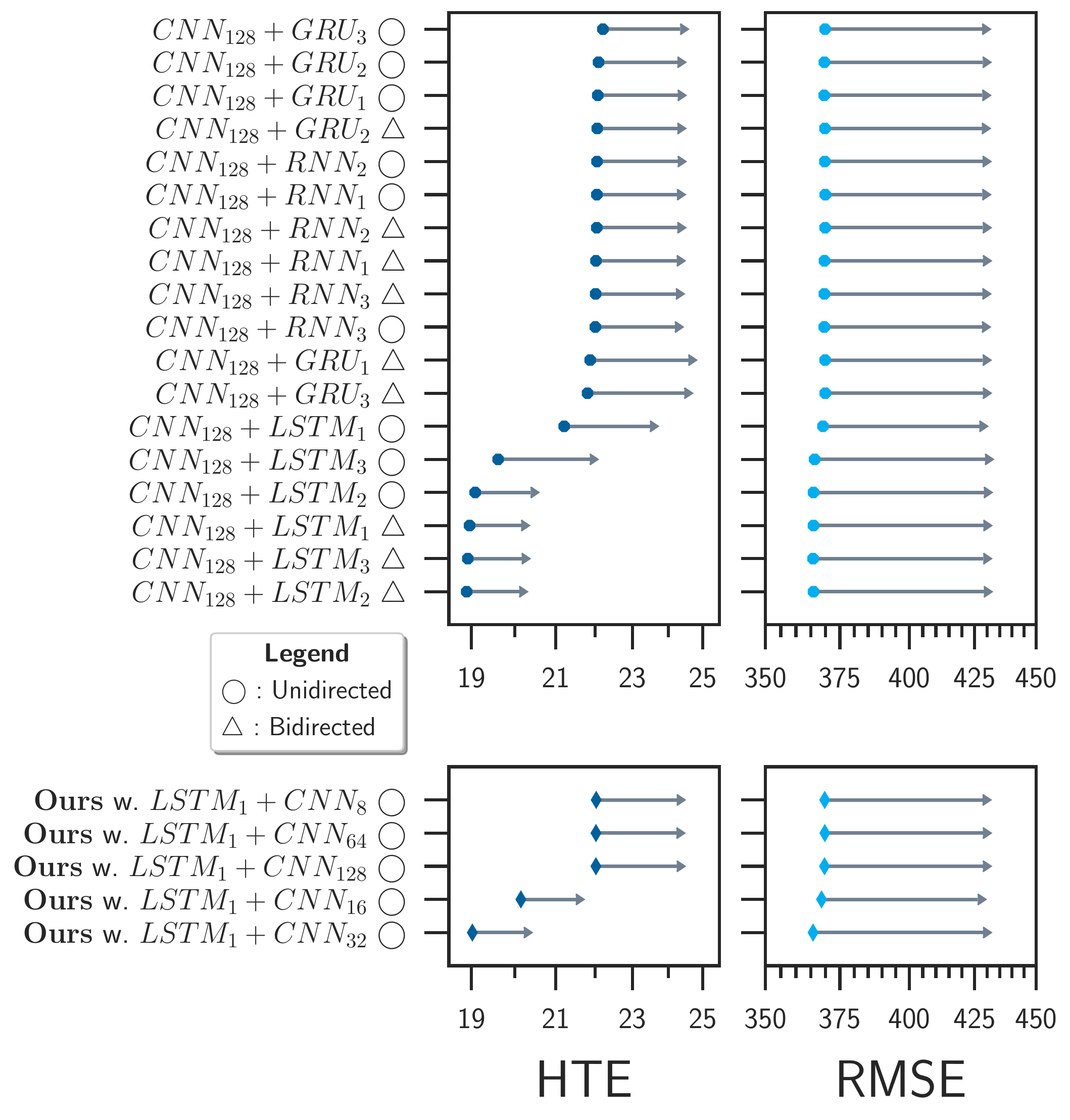}
    \caption{Performance estimation and comparison among different algorithms used for modeling the vessel trajectory network considering the low complexity case where algorithms look for the last $15$ messages to predict the subsequent $25$ messages.
    The performance assessment is based on the Hyperbolic Tangent Error (HTE) and the Root Mean Squared Error (RMSE).
    The experiments were conducted with algorithms on their out-of-the-box version.
    Specifically, among the neural networks, we use $U$ as a superscript to indicate a single-directed model, $B$ for double-directed, and the subscript numbers as the number of stacked recurrent cells.
    The estimators used the same dataset, but the deep learning baselines leveraged our proposed model's HTE loss function and further training adaptation.}
    \label{fig:result-22}
\end{figure}

\begin{figure}[!ht]
    \centering
    \includegraphics[width=\linewidth]{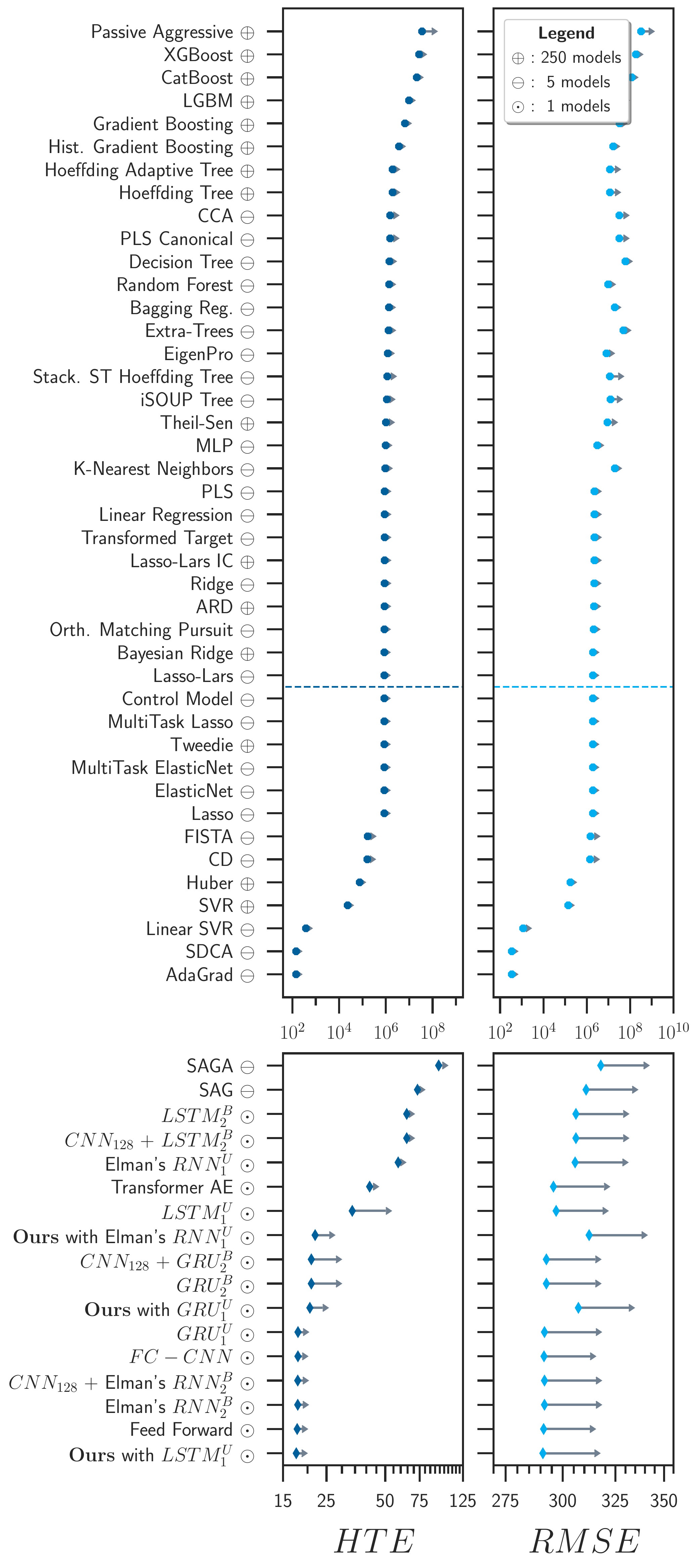}
    \caption{Performance estimation and comparison among different algorithms used for modeling the AIS transmission behavior.
    We have machine and deep learning algorithms clustered in two different segments according to their performance.
    The performance assessment is based on the Hyperbolic Tangent Error (HTE), Mean Absolute Error (MAE), Huber Error (HE), and the Root Mean Squared Error (RMSE).
    The experiments were conducted with algorithms on their out-of-the-box version with no hyperparameter optimization.
    Specifically, Elman's RNN, GRU, and LSTM are bidirectional.
    The estimators used the same dataset, but the deep learning baselines leveraged the HTE loss function and further training adaptation such as the ones used by our proposed model.}
    \label{fig:result-31}
\end{figure}

Figure~\ref{fig:result-31} further supports that adapting the hidden size of the LSTM and the number of input channels of the CNN can improve the performance of the proposed blocks and network architecture, in this case, with more significant improvement in the medium values among the AIS messages, given by the lower HTE, and also the larger values, indicated by an also lower RMSE.
In contrast to Figure~\ref{fig:result-21}, the LSTM-based variations of our model achieve nearly comparable performance.
This indicates LSTMs are more suitable for handling both long and short temporal dependencies of the AIS transmission sequence. This is related to better forecasting sequential AIS message transmission regardless of the presence of outliers in the form of messages too far apart in time and/or space, which can be related to transmission failures or irregular maritime activities (see Section~\ref{sect:introduction}).

\subsubsection*{High complexity case}
Figure~\ref{fig:result-31} presents the performance benchmarks on the prediction of the $50$ subsequent AIS messages given the last $30$ AIS messages observed.
As depicted early, machine learning algorithms are concentrated among the models at the top of the image, and the results presented at the top performed worse than those at the bottom.
As described in Section~\ref{sect:network-architecture}, these models rely on multiple estimators to infer the problem's multiple samples, instants of time, and variables.
These models have a different number of estimators alternating between $5$ to $250$.
In this case, $5$ estimators refer to a different estimator trained per variable of the dataset, while for $250$, we have an estimator per variable and another for each different horizon in the output sequence, holding for Figure~\ref{fig:result-11} and~\ref{fig:result-21}.

\begin{figure}[!b]
    \centering
    \includegraphics[width=.9\linewidth]{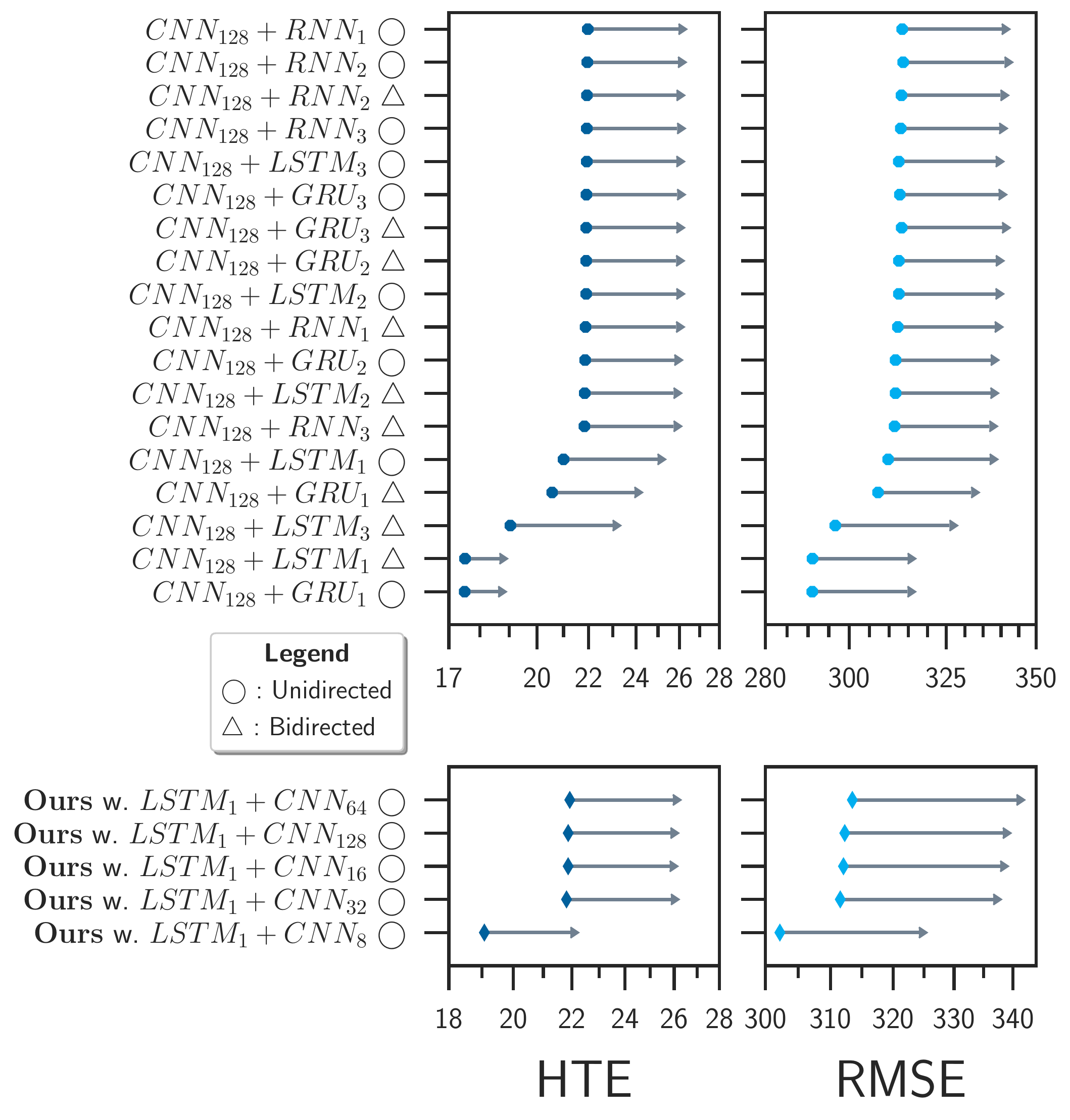}
    \caption{Impact analysis of different Recurrent Neural Networks (RNN) working in different directions and with a varying number of stacked layers compared to our proposed model for modeling the vessel trajectory network.
    In addition, the analysis of the impact of the output channels from the Convolutional Neural Network (CNN) in our proposed modeling approach.
    The performance assessment is based on the Hyperbolic Tangent Error (HTE), Mean Absolute Error (MAE), Huber Error (HE), and the Root Mean Squared Error (RMSE).
    Besides the ones indicated in the image, no other hyperparameter was changed.}
    \label{fig:result-32}
\end{figure}

In particular, {\it Huber} is among those consisting of $250$ different estimators located below the control line.
This is related to the fact that it uses the Huber loss, a smoothed version of the Hyperbolic Tangent using the Mean Absolute Error (MAE) to be less sensitive to outliers.
The same can be observed with the {\it Linear SVR}, which is an ensemble of $5$ estimators and has the MAE with the soft-margin criterion as the loss function to be less sensitive to outliers.
Other relevant-to-mention algorithms are based on linear regressors with different stochastic solvers or optimization mechanisms, such as {\it AdaGrad}, {\it SAG}, and {\it SAGA}.
The reasonable performance of linear-based algorithms comes from the linear nature of consecutive AIS messages, as seen in the low complexity case, which does not incur many variations in the vessel coordinates besides their course and speed over the ground.
A linear estimator can sufficiently model the problem for these particular cases, as an MLP does.
However, when the sequences start to increase, such as in the medium and high complexity cases, the behavior shifts in favor of our approach, showing that the hidden features extracted by the convolutional layer and later processed through the long-short-term memory network can improve the solution.

Through this set of experiments, we observed that the behavior of neural networks diverges significantly according to the predicted sequence's complexity.
Thus, models with fewer non-linearities tend to demonstrate better results for more minor sequences than other more intricate models.
This observation comes from the case of a feed-forward neural network ({\it i.e.}, MLP) being among the top performers for the case of lesser complexity (see Figure~\ref{fig:result-11}), showing better performance than some recurrent neural networks submitted to the same task.
For the high-complexity case only, GRUs showed in Figure~\ref{fig:result-32} to be an alternative for the recurrent unit over large sequences.
That is something to be considered, as the GRU has a simpler formulation than LSTM has and is more efficient and easier to train. Therefore, GRU is a feasible alternative for scaling the proposed architecture and blocks to even larger sequences than used in this work.

\begin{table}[!htb]
    \centering
    \caption{Analysis of the Relative Percentage Difference (RPD) over the three different complexity cases. The results in bold indicate the best-performing ones. Among the algorithms, we included those that consistently performed better than the {\it Control Model}, along with the three complexity case studies.}
    \label{tab:rdp-results}
    \resizebox{\columnwidth}{!}{%
        \begin{tabular}{r|cc|cc|cc|}
            \cline{2-7}
            &
            \multicolumn{6}{c|}{\textit{\textbf{Complexity Level}}} \\ \cline{2-7} 
            &
            \multicolumn{2}{c|}{\textit{{Low}}} &
            \multicolumn{2}{c|}{\textit{{Medium}}} &
            \multicolumn{2}{c|}{\textit{{High}}} \\ \cline{2-7} 
            \textbf{Algorithms} &
            \multicolumn{1}{c|}{\textbf{RPD}} &
            \textbf{+/-} &
            \multicolumn{1}{c|}{\textbf{RPD}} &
            \textbf{+/-} &
            \multicolumn{1}{c|}{\textbf{RPD}} &
            \textbf{+/-} \\ \cline{2-7} 
            AdaGrad & \multicolumn{1}{c|}{1.185} & 0.007 & \multicolumn{1}{c|}{1.202} & 0.012 & \multicolumn{1}{c|}{1.230} & 0.010 \\ \cline{2-7} 
            CD & \multicolumn{1}{c|}{1.407} & 0.008 & \multicolumn{1}{c|}{1.417} & 0.011 & \multicolumn{1}{c|}{1.420} & 0.010 \\ \cline{2-7} 
            $CNN_{128}$ + $GRU^{B}_2$ & \multicolumn{1}{c|}{0.923} & 0.005 & \multicolumn{1}{c|}{0.400} & 0.012 & \multicolumn{1}{c|}{0.440} & 0.100 \\ \cline{2-7} 
            $CNN_{128}$ + $LSTM^{B}_2$ & \multicolumn{1}{c|}{0.382} & 0.010 & \multicolumn{1}{c|}{0.395} & 0.013 & \multicolumn{1}{c|}{1.380} & 0.010 \\ \cline{2-7} 
            $CNN_{128}$ + Elman's $RNN^{B}_2$ & \multicolumn{1}{c|}{0.730} & 0.420 & \multicolumn{1}{c|}{0.392} & 0.008 & \multicolumn{1}{c|}{0.400} & 0.010 \\ \cline{2-7} 
            ElasticNet & \multicolumn{1}{c|}{1.343} & 0.011 & \multicolumn{1}{c|}{1.341} & 0.012 & \multicolumn{1}{c|}{1.340} & 0.010 \\ \cline{2-7} 
            Elman's $RNN^{B}_2$ & \multicolumn{1}{c|}{0.730} & 0.420 & \multicolumn{1}{c|}{0.392} & 0.008 & \multicolumn{1}{c|}{0.400} & 0.010 \\ \cline{2-7} 
            Elman's $RNN^{U}_1$ & \multicolumn{1}{c|}{0.924} & 0.051 & \multicolumn{1}{c|}{0.456} & 0.066 & \multicolumn{1}{c|}{0.960} & 0.010 \\ \cline{2-7} 
            $FC-CNN$ & \multicolumn{1}{c|}{0.401} & 0.022 & \multicolumn{1}{c|}{0.403} & 0.005 & \multicolumn{1}{c|}{0.440} & 0.060 \\ \cline{2-7} 
            Feed Forward & \multicolumn{1}{c|}{0.383} & 0.013 & \multicolumn{1}{c|}{0.407} & 0.021 & \multicolumn{1}{c|}{0.410} & 0.010 \\ \cline{2-7} 
            FISTA & \multicolumn{1}{c|}{1.409} & 0.008 & \multicolumn{1}{c|}{1.418} & 0.011 & \multicolumn{1}{c|}{1.420} & 0.010 \\ \cline{2-7} 
            $GRU^{B}_2$ & \multicolumn{1}{c|}{0.923} & 0.005 & \multicolumn{1}{c|}{0.400} & 0.012 & \multicolumn{1}{c|}{0.440} & 0.100 \\ \cline{2-7} 
            $GRU^{U}_1$ & \multicolumn{1}{c|}{0.514} & 0.176 & \multicolumn{1}{c|}{0.525} & 0.121 & \multicolumn{1}{c|}{0.400} & 0.010 \\ \cline{2-7} 
            Huber & \multicolumn{1}{c|}{1.336} & 0.011 & \multicolumn{1}{c|}{1.334} & 0.010 & \multicolumn{1}{c|}{1.330} & 0.020 \\ \cline{2-7} 
            Lasso & \multicolumn{1}{c|}{1.343} & 0.011 & \multicolumn{1}{c|}{1.341} & 0.012 & \multicolumn{1}{c|}{1.340} & 0.010 \\ \cline{2-7} 
            Linear SVR & \multicolumn{1}{c|}{1.297} & 0.018 & \multicolumn{1}{c|}{1.310} & 0.009 & \multicolumn{1}{c|}{1.350} & 0.010 \\ \cline{2-7} 
            $LSTM^{B}_2$ & \multicolumn{1}{c|}{0.382} & 0.010 & \multicolumn{1}{c|}{0.395} & 0.013 & \multicolumn{1}{c|}{1.380} & 0.010 \\ \cline{2-7} 
            $LSTM^{U}_1$ & \multicolumn{1}{c|}{1.293} & 0.194 & \multicolumn{1}{c|}{0.988} & 0.364 & \multicolumn{1}{c|}{0.610} & 0.220 \\ \cline{2-7} 
            MultiTask ElasticNet & \multicolumn{1}{c|}{1.343} & 0.011 & \multicolumn{1}{c|}{1.341} & 0.012 & \multicolumn{1}{c|}{1.340} & 0.010 \\ \cline{2-7} 
            MultiTask Lasso & \multicolumn{1}{c|}{1.343} & 0.011 & \multicolumn{1}{c|}{1.341} & 0.012 & \multicolumn{1}{c|}{1.340} & 0.010 \\ \cline{2-7} 
            ${\bf Ours}$~w/~$GRU^{U}_1$ & \multicolumn{1}{c|}{\textbf{0.365}} & 0.011 & \multicolumn{1}{c|}{0.399} & 0.019 & \multicolumn{1}{c|}{0.410} & 0.020 \\ \cline{2-7} 
            ${\bf Ours}$~w/~$LSTM^{U}_1$ & \multicolumn{1}{c|}{0.368} & 0.011 & \multicolumn{1}{c|}{\textbf{0.376}} & 0.009 & \multicolumn{1}{c|}{\textbf{0.380}} & 0.010 \\ \cline{2-7} 
            ${\bf Ours}$~w/~Elman's $RNN^{U}_1$ & \multicolumn{1}{c|}{0.376} & 0.020 & \multicolumn{1}{c|}{0.397} & 0.010 & \multicolumn{1}{c|}{0.410} & 0.010 \\ \cline{2-7} 
            SAG & \multicolumn{1}{c|}{0.835} & 0.009 & \multicolumn{1}{c|}{0.840} & 0.015 & \multicolumn{1}{c|}{0.840} & 0.020 \\ \cline{2-7} 
            SAGA & \multicolumn{1}{c|}{0.962} & 0.030 & \multicolumn{1}{c|}{1.000} & 0.031 & \multicolumn{1}{c|}{1.040} & 0.060 \\ \cline{2-7} 
            SDCA & \multicolumn{1}{c|}{1.185} & 0.007 & \multicolumn{1}{c|}{1.202} & 0.012 & \multicolumn{1}{c|}{1.230} & 0.010 \\ \cline{2-7} 
            SVR & \multicolumn{1}{c|}{1.342} & 0.011 & \multicolumn{1}{c|}{1.337} & 0.013 & \multicolumn{1}{c|}{1.330} & 0.010 \\ \cline{2-7} 
            Transformer AE & \multicolumn{1}{c|}{0.752} & 0.012 & \multicolumn{1}{c|}{0.752} & 0.015 & \multicolumn{1}{c|}{0.750} & 0.010 \\ \cline{2-7} 
            Tweedie & \multicolumn{1}{c|}{1.343} & 0.011 & \multicolumn{1}{c|}{1.341} & 0.012 & \multicolumn{1}{c|}{1.340} & 0.010 \\ \cline{2-7} 
        \end{tabular}%
    }
\end{table}

\subsection*{Results Interpretability}
Due to the narrow interpretation of the HTE and RMSE, Table~\ref{tab:rdp-results} shows the Relative Percentage Difference (RPD) results.
Such a metric evaluates how far the forecasted message is from the expected message.
As the results of the RPD can be both positive and negative, we can understand if the predictions are lower or higher than the expected value.
For the RPD formulation, the results can be higher than $100\%$, meaning that the error can be multiple times larger than the expected value.
In this sense, reasonable results are below $50\%$ and the closer to $0\%$ ({\it i.e.}, perfect model), the better.

The RPD results show a different behavior from the previous metrics, where our proposal consistently shows greater stability in the shared error of forecasting the AIS messages.
The models that previously showed great efficiency now show slightly worse results.
For experiments with short, medium, and large-sized AIS messages sequences, our model achieved $36/37/38\%$ of the RPD, while Elman's RNN scored $92/45/96\%$, GRU scored $51/52/40\%$, and LSTM scored $129/98/61\%$.
This means that the proposed solution showed greater performance in forecasting the content of the AIS message, including the vessel positioning and other dynamic variables such as COG, SOG, and delta time of consecutive messages.
This is not only important for controlling and increasing awareness about the AIS transmission system, but it has the potential to be used in detecting misleading transmission patterns, such as on-off AIS transceiver behavior modeling and AIS spoofing activity detection.
The variation from the results observed in the HTE regarding the RPD is due to the non-linear nature of the Hyperbolic Tangent, which might not show the same ability as previously observed when in a linear space.
That leads us to conclude that our modeling solution over-performs the competing models in all three complexity cases, being more robust to irregular timing.

\subsection*{Model Ablation}
Lastly, Table~\ref{tab:further-results} presents the ablation results, highlighting how the traditional LSTM, the FC-CNN, the LSTM-CNN single block, and the LSTM-CNN-AR in a double-block structure behave according to the RPD.
Through these results, we observe that our single-block architecture shows suitable performance in all three cases.
However, it can leverage the further performance of an additional block in the low complexity case, which relates to the improved performance of stacked RNNs observed when describing the low-complexity case.
The fully connected convolutional layer alone has not shown a favorable result compared to the others, but it outperformed the traditional LSTM also in the three scenarios.
Overall, the experiments support the proposed modeling approach, demonstrating effectiveness on different horizon sizes.

\begin{table}[!htb]
    \centering
    \caption{Detailed results for the proposed modeling approach and further network components describing the Relative Percentage Difference (RPD) and the observed standard deviation.}
    \label{tab:further-results}
    % \resizebox{.9\columnwidth}{!}{%
        \begin{tabular}{lcc}
            \hline
            \multicolumn{1}{|l|}{Low Complexity} & \multicolumn{1}{c|}{\textbf{RPD}} & \multicolumn{1}{c|}{(+/-)} \\ \hline
            $LSTM^{U}_1$ & 1.2933 & 0.1941 \\ \hline
            $FC-CNN_{128}$ & 0.4009 & 0.0222 \\ \hline
            \textbf{Single Block:} $LSTM^{U}_1$ w. $CNN_{128}$ & 0.368 & 0.011 \\ \hline
            \textbf{Double Block:} $LSTM^{U}_1$ w. $CNN_{16}$ + AR & \textbf{0.356} & 0.012 \\ \hline
            \multicolumn{1}{|l|}{Medium Complexity} & \multicolumn{1}{c|}{HTE} & \multicolumn{1}{c|}{(+/-)} \\ \hline
            $LSTM^{U}_1$ & 0.9883 & 0.3643 \\ \hline
            $FC-CNN_{128}$ & 0.4031 & 0.0053 \\ \hline
            \textbf{Single Block:} $LSTM^{U}_1$ w. $CNN_{128}$ & 0.376 & 0.008 \\ \hline
            \textbf{Double Block:} $LSTM^{U}_1$ w. $CNN_{32}$ + AR & \textbf{0.374} & 0.012 \\ \hline
            \multicolumn{1}{|l|}{High Complexity} & \multicolumn{1}{c|}{HTE} & \multicolumn{1}{c|}{(+/-)} \\ \hline
            $LSTM^{U}_1$ & 0.6083 & 0.22 \\ \hline
            $FC-CNN_{128}$ & 0.4392 & 0.06 \\ \hline
            \textbf{Single Block:} $LSTM^{U}_1$ w. $CNN_{128}$ & \textbf{0.383} & 0.010 \\ \hline
            \textbf{Double Block:} $LSTM^{U}_1$ w. $CNN_{8}$ + AR & 0.395 & 0.015 \\ \hline
        \end{tabular}%
    % }
\end{table}

\subsection*{Limitations}
{
Due to working with multiple trajectories simultaneously, we provide additional information concerning the transmission behavior of AIS messages.
However, it also turns the problem into a more significant challenge for the models due to the increased uncertainty related to the irregular timing of messages.
The temporal irregularity between consecutive transmitted AIS messages is considered to be noise, and it turns the AIS messages into outliers when the gap between two messages is too large.
By working with multiple trajectories, their presence is even more significant.
This issue would be reduced if working with smoothed trajectories because they include virtual AIS messages to fill the temporal gaps and interpolate the trajectory.
However, that does not mean that the trajectories will be equally accurately pictured once interpolated due to not being free of uncertainty when the temporal gap is too large.
Also, interpolating every trajectory of the dataset might not be straightforward in near real-time conditions such as observed in AIS data streams.

Our proposal shows a different perspective on dealing with this problem.
The significant difference is that interpolation techniques are preprocessed prior to the analysis.
However, our approach works on cases where that does not hold, {\it i.e.}, on the raw data.
As such, we transfer the responsibility of smoothing the trajectories and reducing the irregularities by randomly inputting increased amounts of temporal data and guiding the algorithm to avoid pitfalls related to the outlier messages.
While this may not be the most straightforward approach due to the complexity of training the network, it has been shown to perform better according to the experiments.

It is evident that generalization and specification are opposite qualities of a learning model.
That being said, our model behaves and generalizes better over multiple trajectories simultaneously.
However, when a trajectory of a single vessel is of interest and the historical AIS data from the vessel of interest is available, a model focused on the specific vessel might yield better forecasting results over its trajectory.
That is because models trained for forecasting the trajectory of a single vessel on the observed data of the vessel of interest will capture the particular behavior of that vessel.
Regardless, our modeling approach showed more performance and robustness than other modeling possibilities on the task of simultaneously predicting multiple trajectories on the raw AIS data transmitted along with the vessel's trajectory.

Lastly, we use delta time to include a notion of temporality in the data, but more features are needed to achieve superior performance in this task. Information related to the period of the day and the year's season might allow for a more refined understanding of the transmission patterns, which are closely tied to vessels' mobility patterns correlated to these variables. The same holds for geophysical data, such as information about the winds, the waves, tidal patterns, and the weather, which have the potential to refine this process further because the mobility pattern is also expected to change under harsh navigation conditions. The data fusion not covered in our study seems to show potential to further studies in this area.
}

\section{Conclusions}
\label{sect:conclusions}

This paper addresses modeling the AIS message transmission behavior through neural networks under noisy and temporally irregular data.
We presented a comprehensive set of experiments comprising multiple machine and deep learning algorithms submitted to forecasting tasks with horizon sizes of varying lengths.
Such results show that traditional machine learning models strive to generalize over many vessels.
Deep learning models revealed themselves to easily capture the temporal irregularity while preserving the spatial awareness when forecasting the trajectories of different vessels, given the lower Relative Percentage Error (RPD) assessed on three different complexity cases.
The models showed to be more robust to the AIS messages' temporal irregularity and delivered beneficial results over machine learning algorithms, mainly when combined with convolutional layers.

More specifically, joining long-short-term memory neural networks with single-dimension convolutional neural networks enhances the feature extraction process, increasing the neural network's performance under different circumstances.
The results show that our model improves the prediction of vessel routes when analyzing multiple vessels of diverging types simultaneously.
This translates into a model that, on average, provides more accurate forecasting results over multiple trajectories rather than a model tailored for a single class of vessels or trained on long historical sequences of AIS messages of a single vessel.
In such a case, deep learning models achieve better results than competing algorithms, mainly when joining convolutional and recurrent networks.

Experimenting with short, medium, and large-sized AIS messages sequences, the proposed model achieved $36/37/38\%$ of the RPD, whereas we observed $92/45/96\%$ on the Elman's RNN, $51/52/40\%$ on the GRU, and $129/98/61\%$ on the LSTM network.
Besides the performance improvement derived from our alternative network architecture, we also observed that our model was more numerically stable over the experiments using different window and horizon sizes, showing better performance in forecasting both short and long AIS message sequences simultaneously for multiple vessels of different types.
Through such a multifaceted analysis of estimators' performance, we concluded that our modeling approach performs better on different sizes of AIS sequences.
It also allows further improvement by adapting the numbers of output channels of the convolution feature-extraction layer, which can increase or decrease the number of temporal samples the model will use for training.

Nevertheless, much improvement can be achieved along with similar study premises.
Those would be related to increasing the geographical boundary of AIS messages to a global scale, which would require greater computational power and processing time.
Further improvement refers to using different modeling approaches for the AIS message data, such as motif analysis on grided AIS data.
Additionally, different neural network techniques could enhance the interaction between trajectories.
This is the case of Graph Neural Networks (GNNs), which might shape the relationship of the variables within the AIS messages, and network embeddings that can be used to bring further knowledge about the mobility of the vessel to the forecasting pipeline.

\section*{Authors contributions statement}
Conceptualization, G.S. and M.F.; methodology, G.S. and M.F.; software, G.S. and M.F.; validation, G.S., M.F., A.S., and S.M.; formal analysis, G.S. and M.F.; investigation, G.S. and M.F.; resources, S.M.; data curation, G.S., M.D., and A.S.; writing ---original draft preparation, G.S. and M.F.; writing --- review and editing, M.F., G.S., A.S., and S.M.; visualization, G.S. and M.F.; supervision, S.M.; project administration, S.M.; funding acquisition, S.M. All authors have read and agreed to the published version of the manuscript.

\section*{Acknowledgments}
The authors thank {\it Spire} (former {\it exactEarth}) for the vessel trajectory network dataset, {\it M. Smith} for assisting in the data extraction, and {\it M. Gillis} for reviewing the final manuscript.
This research was partially funded by the Institute for Big Data Analytics (IBDA) and the Ocean Frontier Institute (OFI) at Dalhousie University, Halifax - NS, Canada; and further funded by the Canada First Research Excellence Fund (CFREF), the Canadian Foundation for Innovation MERIDIAN cyberinfrastructure\footnote{\url{https://meridian.cs.dal.ca/}}, and the Natural Sciences and Engineering Research Council of Canada (NSERC).

% Loading bibliography style file
\bibliographystyle{unsrt}

\bibliography{3-references}

% \newpage
\vspace{-1.25cm}
\begin{IEEEbiography}[{\includegraphics[width=1in, height=1.25in, clip, keepaspectratio]{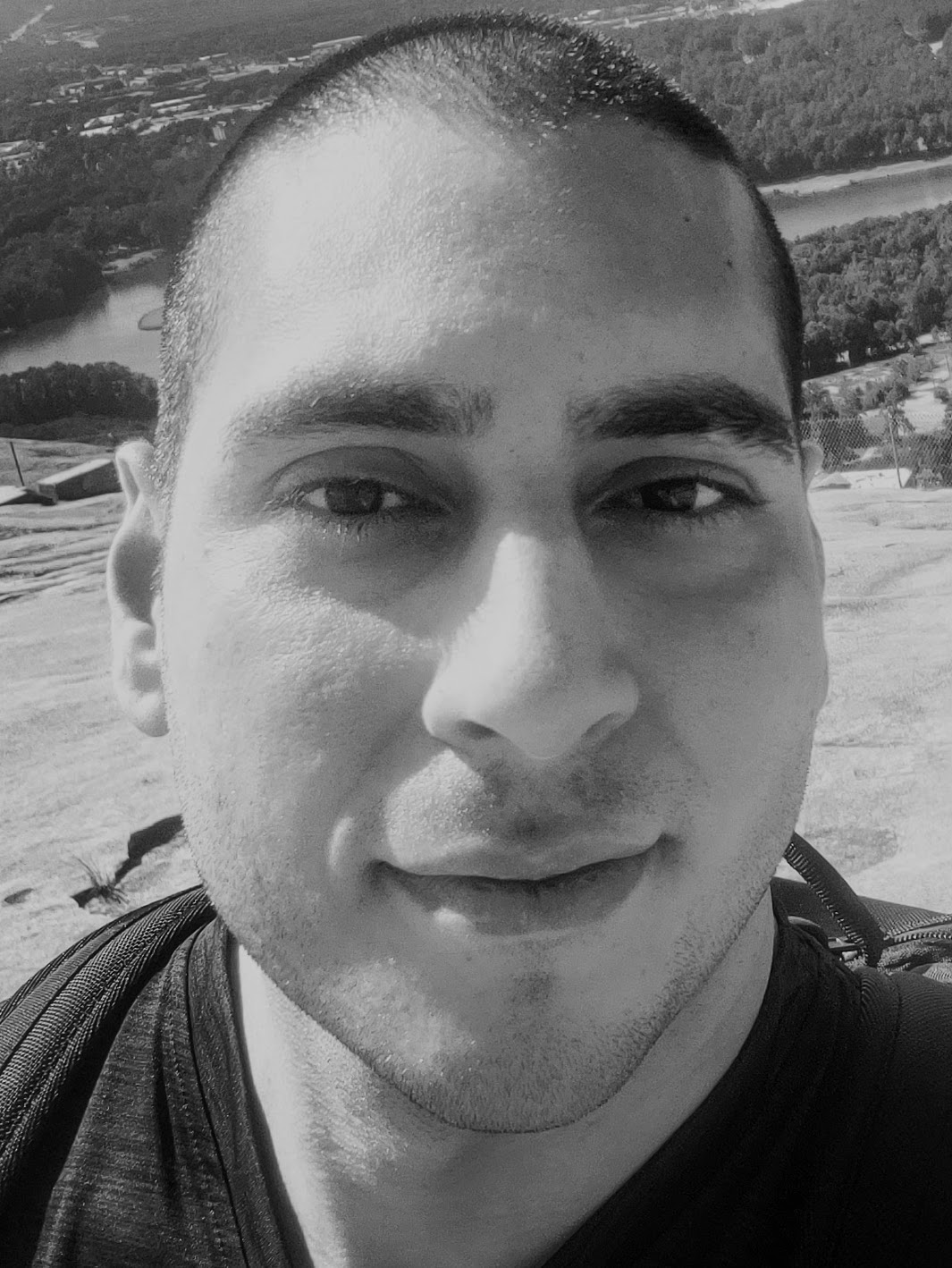}}]{Gabriel Spadon} is currently a postdoctoral fellow at Dalhousie University, Canada, working on projects related to vessel mobility and underwater acoustics to architect neural network for improving ocean awareness and monitoring capabilities. He has a PhD (with honors) in Computer Science at the University of Sao Paulo, Brazil, part of which was carried out with the Georgia Institute of Technology, USA. Spadon has worked intensively on network science and artificial intelligence during the last few years. He has authored (and co-authored) several research articles on knowledge discovery through complex networks and data mining. His current research interests include neural-inspired models, graph-based learning, and complex networks.
\end{IEEEbiography}
% ------------ %
\vspace{-1cm} % bug-fix
% ------------ %
\begin{IEEEbiography}[{\includegraphics[width=1in, height=1.25in, clip, keepaspectratio]{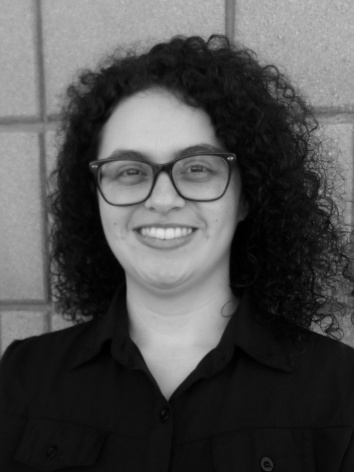}}]{Martha D. Ferreira} is currently a postdoctoral fellow at Dalhousie University, working on a project with GDMS-C and DRDC to evaluate and develop an approach in the context of suspicious or dangerous activities in the physical marine environment. She got her Ph.D. at the University of Sao Paulo, Sao Carlos, in March 2019. The research was in the Deep Learning area, focusing on Convolutional Neural Networks, including a formalization of CNN aspects and Statistical Learning Theory to prove CNNs generalization. Her research interests are Machine Learning, Deep Learning, Time Series Analysis, and Information Retrieval.
\end{IEEEbiography}

\begin{IEEEbiography}[{\includegraphics[width=1in, height=1.25in, clip, keepaspectratio]{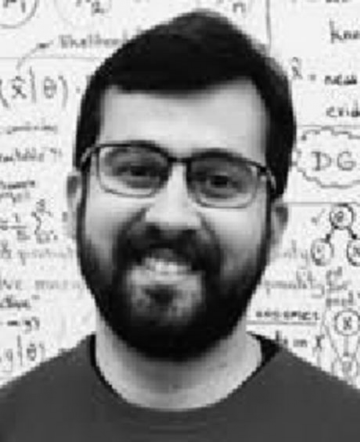}}]{Amilcar Soares} is currently an Assistant Professor at the Memorial University of Newfoundland at the Department of Computer Science. His research interests include spatiotemporal data segmentation, classification, enrichment, and visualization. He holds a Ph.D. in computer science from the Federal University of Pernambuco. He has been involved in several research projects funded by the Natural Sciences and Engineering Research Council of Canada (NSERC), Department of Fisheries and Oceans (DFO), Transport Canada (TC), and Defence Research and Development Canada (DRDC).
\end{IEEEbiography}
\pagebreak

\begin{IEEEbiography}[{\includegraphics[width=1in, height=1.25in, clip, keepaspectratio]{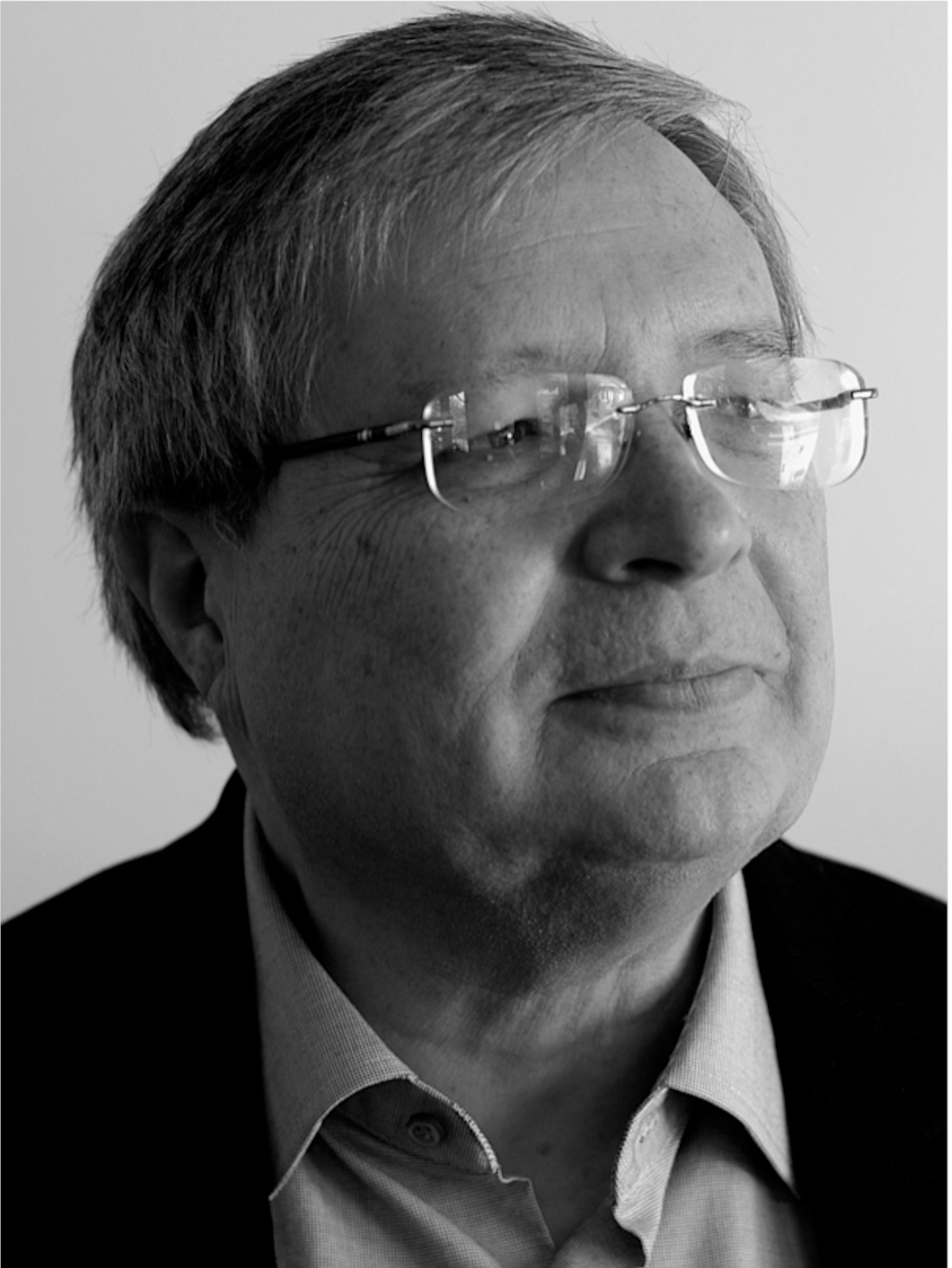}}]{Stan Matwin} is currently the director of the Institute for Big Data Analytics, Dalhousie University, Halifax, Nova Scotia, where he is a professor and Canada Research Chair (Tier 1) in Interpretability for Machine Learning. He is also a distinguished (Emeritus) professor at the University of Ottawa and a full professor at the Institute of Computer Science, Polish Academy of Sciences. His main research interests include big data, text mining, machine learning, and data privacy. He is a member of the Editorial Boards of \textit{IEEE Transactions on Knowledge and Data Engineering} and \textit{Journal of Intelligent Information Systems}. He received the Lifetime Achievement Award from the Canadian AI Association (CAIAC).
\end{IEEEbiography}

\EOD % LaTeX EOF
\end{document}

\let\WriteBookmarks\relax
\def\textpagefraction{.001}
\def\floatpagepagefraction{1}

\title{{\bf Supplementary Information}\\\vspace{.3cm}
Unfolding AIS transmission behavior for vessel movement modeling on noisy data leveraging machine learning}
\author{Gabriel Spadon, Martha D. Ferreira, Amilcar Soares, Stan Matwin}
\date{} % removing the date
\maketitle

\section*{List of Machine Learning Algorithms}

\begin{longtable}[c]{r|r|l|l}
	\hline
	\# & \textbf{Acronym}         & \textbf{Name}                                        & \textbf{Source}              \\ \hline
	\endhead
	1  & ARD                      & Automatic Relevance Determination Regression         & \url{https://bit.ly/3146kQA} \\ \hline
	2  & AdaBoost                 & ---                                                  & \url{https://bit.ly/3qyfpd1} \\ \hline
	3  & AdaGrad                  & ---                                                  & \url{https://bit.ly/3HlkITV} \\ \hline
	4  & Adap. Random Forest      & Adaptive Random Forest                               & \url{https://bit.ly/32Fpywy} \\ \hline
	5  & Bagging Reg.             & Bagging Regressor                                    & \url{https://bit.ly/3Jo70lf} \\ \hline
	6  & Bayesian Ridge           & ---                                                  & \url{https://bit.ly/3pxIINC} \\ \hline
	7  & CCA                      & Canonical Correlation Analysis                       & \url{https://bit.ly/3FE25KO} \\ \hline
	8  & CD                       & Coordinate Descent                                   & \url{https://bit.ly/3Hj8UBG} \\ \hline
	9  & CatBoost                 & ---                                                  & \url{https://bit.ly/3sFxE2T} \\ \hline
	10 & Decision Tree            & ---                                                  & \url{https://bit.ly/33XoDbf} \\ \hline
	11 & Control Model            & Control Model                                        & \url{https://bit.ly/3FAb6EH} \\ \hline
	12 & EigenPro                 & \cellcolor[HTML]{FFFFFF}---                          & \url{https://bit.ly/3pwhl6M} \\ \hline
	13 & ElasticNet               & ---                                                  & \url{https://bit.ly/3mD43DD} \\ \hline
	14 & Extra-Trees              & ---                                                  & \url{https://bit.ly/3qzt97H} \\ \hline
	15 & Factorization Machine    & ---                                                  & \url{https://bit.ly/3qyxvvs} \\ \hline
	16 & FISTA                    & FISTA                                                & \url{https://bit.ly/3puWRv3} \\ \hline
	17 & Gamma                    & Generalized Linear Model with a Gamma distribution   & \url{https://bit.ly/32wofjI} \\ \hline
	18 & Gaussian Process         & Gaussian Process Regression                          & \url{https://bit.ly/3zdvRUg} \\ \hline
	19 & Gradient Boosting        & ---                                                  & \url{https://bit.ly/3186FSs} \\ \hline
	20 & Hist. Gradient Boosting  & Histogram-based Gradient Boosting Regression Tree    & \url{https://bit.ly/3FuXt9D} \\ \hline
	21 & Hoeffding Adaptive Tree  & ---                                                  & \url{https://bit.ly/348avMF} \\ \hline
	22 & Hoeffding Tree           & ---                                                  & \url{https://bit.ly/3sFheaM} \\ \hline
	23 & Huber                    & ---                                                  & \url{https://bit.ly/3sJpqqD} \\ \hline
	24 & K-Nearest Neighbors      & ---                                                  & \url{https://bit.ly/3186JSc} \\ \hline
	25 & Kernel Ridge             & ---                                                  & \url{https://bit.ly/3HmsD3F} \\ \hline
	26 & LGBM                     & Light Gradient Boosting Machine                      & \url{https://bit.ly/3HnTeNP} \\ \hline
	27 & Lars                     & Least Angle Regression                               & \url{https://bit.ly/3z5PywX} \\ \hline
	28 & Lasso                    & ---                                                  & \url{https://bit.ly/32GkYOq} \\ \hline
	29 & Lasso-Lars               & ---                                                  & \url{https://bit.ly/3qyvFe2} \\ \hline
	30 & Lasso-Lars IC            & Lasso-Lars with Information Criterion                & \url{https://bit.ly/3HlqLrA} \\ \hline
	31 & Linear Regression        & ---                                                  & \url{https://bit.ly/3FMuKgs} \\ \hline
	32 & Linear SVR               & Linear Support Vector Regression                     & \url{https://bit.ly/3EEvnY5} \\ \hline
	33 & MLP                      & Multi-layer Perceptron Regressor                     & \url{https://bit.ly/3ewXJJi} \\ \hline
	34 & MultiTask ElasticNet     & ---                                                  & \url{https://bit.ly/3H91hOi} \\ \hline
	35 & MultiTask Lasso          & ---                                                  & \url{https://bit.ly/3qO98uf} \\ \hline
	36 & NuSVR                    & Nu Support Vector Regression                         & \url{https://bit.ly/3FCPQOn} \\ \hline
	37 & Orth. Matching Pursuit   & Orthogonal Matching Pursuit                          & \url{https://bit.ly/3JsaWl8} \\ \hline
	38 & PLS Canonical            & Partial Least Square Canonical                       & \url{https://bit.ly/3sJpUgr} \\ \hline
	39 & PLS                      & Partial Least Squares                                & \url{https://bit.ly/3JrNXGz} \\ \hline
	40 & Passive Aggressive       & ---                                                  & \url{https://bit.ly/3mFM9A1} \\ \hline
	41 & Poisson                  & Generalized Linear Model with a Poisson distribution & \url{https://bit.ly/3Hmyvdb} \\ \hline
	42 & Polynomial Network       & ---                                                  & \url{https://bit.ly/3pwjX4A} \\ \hline
	43 & Quantile                 & ---                                                  & \url{https://bit.ly/3Jn6xQc} \\ \hline
	44 & Radius Neighbors         & ---                                                  & \url{https://bit.ly/32HYHj7} \\ \hline
	45 & Random Forest            & ---                                                  & \url{https://bit.ly/32tU78B} \\ \hline
	46 & Ridge                    & ---                                                  & \url{https://bit.ly/3sDAsO8} \\ \hline
	47 & SAGA                     & Stochastic Average Gradient Ascent                   & \url{https://bit.ly/3mDzYDW} \\ \hline
	48 & SAG                      & Stochastic Average Gradient                          & \url{https://bit.ly/3quUt6R} \\ \hline
	49 & SDCA                     & Stochastic Dual Coordinate Ascent                    & \url{https://bit.ly/32tUa4h} \\ \hline
	50 & SGD                      & Stochastic Gradient Descent                          & \url{https://bit.ly/32zCvrP} \\ \hline
	51 & SVR                      & Support Vector Regression                            & \url{https://bit.ly/3pAKldK} \\ \hline
	52 & SVRG                     & Stochastic Variance Reduced Gradient                 & \url{https://bit.ly/3HjBfI2} \\ \hline
	53 & Stack. ST Hoeffding Tree & Stacked Single-target Hoeffding Tree                 & \url{https://bit.ly/3qyIaGL} \\ \hline
	54 & Theil-Sen                & ---                                                  & \url{https://bit.ly/345f8ab} \\ \hline
	55 & Transformed Target       & ---                                                  & \url{https://bit.ly/3EvYFbr} \\ \hline
	56 & Tweedie                  & ---                                                  & \url{https://bit.ly/3esTZbO} \\ \hline
	57 & XGBoost                  & ---                                                  & \url{https://bit.ly/3z45dNr} \\ \hline
	58 & iSOUP Tree               & Incremental Structured Output Prediction Tree        & \url{https://bit.ly/3pHuFp7} \\ \hline
    \caption{List of machine learning algorithms specifying their acronym, full name, and availability.}
\end{longtable}

\FloatBarrier%
\clearpage\newpage%
\begin{titlepage}
    \vspace*{\fill}
        \begin{center}
          \Huge{\bf Type of Algorithms \\\rule[5px]{2cm}{.1cm}~and~\rule[5px]{2cm}{.1cm}\\ Usage on Experimentation}
        \end{center}
    \vspace*{\fill}
\end{titlepage}

\begin{longtable}[c]{c|r|c|c|c|c}
	\hline
	\# & \textbf{Acronym}         & \textbf{Type} & \textbf{15-05} & \textbf{15-25} & \textbf{30-50} \\ \hline
	\endhead
	1  & ARD                      & \faBullseye   & \faCheckCircle & \faCheckCircle & \faCheckCircle \\ \hline
	2  & AdaBoost                 & \faBullseye   & \faTimesCircle & \faTimesCircle & \faTimesCircle \\ \hline
	3  & AdaGrad                  & \faDotCircleO & \faCheckCircle & \faCheckCircle & \faCheckCircle \\ \hline
	4  & Adap. Random Forest      & \faBullseye   & \faTimesCircle & \faTimesCircle & \faTimesCircle \\ \hline
	5  & Bagging Reg.             & \faDotCircleO & \faCheckCircle & \faCheckCircle & \faCheckCircle \\ \hline
	6  & Bayesian Ridge           & \faBullseye   & \faCheckCircle & \faCheckCircle & \faCheckCircle \\ \hline
	7  & CCA                      & \faDotCircleO & \faCheckCircle & \faCheckCircle & \faCheckCircle \\ \hline
	8  & CD                       & \faDotCircleO & \faCheckCircle & \faCheckCircle & \faCheckCircle \\ \hline
	9  & CatBoost                 & \faBullseye   & \faCheckCircle & \faCheckCircle & \faCheckCircle \\ \hline
	10 & Decision Tree            & \faDotCircleO & \faCheckCircle & \faCheckCircle & \faCheckCircle \\ \hline
	11 & Control Model            & \faDotCircleO & \faCheckCircle & \faCheckCircle & \faCheckCircle \\ \hline
	12 & EigenPro                 & \faDotCircleO & \faCheckCircle & \faCheckCircle & \faCheckCircle \\ \hline
	13 & ElasticNet               & \faDotCircleO & \faCheckCircle & \faCheckCircle & \faCheckCircle \\ \hline
	14 & Extra-Trees              & \faDotCircleO & \faCheckCircle & \faCheckCircle & \faCheckCircle \\ \hline
	15 & Factorization Machine    & \faBullseye   & \faCheckCircle & \faTimesCircle & \faTimesCircle \\ \hline
	16 & FISTA                    & \faDotCircleO & \faCheckCircle & \faCheckCircle & \faCheckCircle \\ \hline
	17 & Gamma                    & \faBullseye   & \faTimesCircle & \faTimesCircle & \faTimesCircle \\ \hline
	18 & Gaussian Process         & \faDotCircleO & \faTimesCircle & \faTimesCircle & \faTimesCircle \\ \hline
	19 & Gradient Boosting        & \faBullseye   & \faCheckCircle & \faCheckCircle & \faCheckCircle \\ \hline
	20 & Hist. Gradient Boosting  & \faBullseye   & \faCheckCircle & \faCheckCircle & \faCheckCircle \\ \hline
	21 & Hoeffding Adaptive Tree  & \faBullseye   & \faCheckCircle & \faCheckCircle & \faCheckCircle \\ \hline
	22 & Hoeffding Tree           & \faBullseye   & \faCheckCircle & \faCheckCircle & \faCheckCircle \\ \hline
	23 & Huber                    & \faBullseye   & \faCheckCircle & \faCheckCircle & \faCheckCircle \\ \hline
	24 & K-Nearest Neighbors      & \faDotCircleO & \faCheckCircle & \faCheckCircle & \faCheckCircle \\ \hline
	25 & Kernel Ridge             & \faDotCircleO & \faTimesCircle & \faTimesCircle & \faTimesCircle \\ \hline
	26 & LGBM                     & \faBullseye   & \faCheckCircle & \faCheckCircle & \faCheckCircle \\ \hline
	27 & Lars                     & \faDotCircleO & \faCheckCircle & \faCheckCircle & \faTimesCircle \\ \hline
	28 & Lasso                    & \faDotCircleO & \faCheckCircle & \faCheckCircle & \faCheckCircle \\ \hline
	29 & Lasso-Lars               & \faDotCircleO & \faCheckCircle & \faCheckCircle & \faCheckCircle \\ \hline
	30 & Lasso-Lars IC            & \faBullseye   & \faCheckCircle & \faCheckCircle & \faCheckCircle \\ \hline
	31 & Linear Regression        & \faDotCircleO & \faCheckCircle & \faCheckCircle & \faCheckCircle \\ \hline
	32 & Linear SVR               & \faDotCircleO & \faCheckCircle & \faCheckCircle & \faCheckCircle \\ \hline
	33 & MLP                      & \faDotCircleO & \faCheckCircle & \faCheckCircle & \faCheckCircle \\ \hline
	34 & MultiTask ElasticNet     & \faDotCircleO & \faCheckCircle & \faCheckCircle & \faCheckCircle \\ \hline
	35 & MultiTask Lasso          & \faDotCircleO & \faCheckCircle & \faCheckCircle & \faCheckCircle \\ \hline
	36 & NuSVR                    & \faBullseye   & \faTimesCircle & \faTimesCircle & \faTimesCircle \\ \hline
	37 & Orth. Matching Pursuit   & \faDotCircleO & \faCheckCircle & \faCheckCircle & \faCheckCircle \\ \hline
	38 & PLS Canonical            & \faDotCircleO & \faCheckCircle & \faCheckCircle & \faCheckCircle \\ \hline
	39 & PLS                      & \faDotCircleO & \faCheckCircle & \faCheckCircle & \faCheckCircle \\ \hline
	40 & Passive Aggressive       & \faBullseye   & \faCheckCircle & \faCheckCircle & \faTimesCircle \\ \hline
	41 & Poisson                  & \faBullseye   & \faTimesCircle & \faTimesCircle & \faTimesCircle \\ \hline
	42 & Polynomial Network       & \faBullseye   & \faCheckCircle & \faTimesCircle & \faTimesCircle \\ \hline
	43 & Quantile                 & \faBullseye   & \faTimesCircle & \faTimesCircle & \faTimesCircle \\ \hline
	44 & Radius Neighbors         & \faDotCircleO & \faTimesCircle & \faTimesCircle & \faTimesCircle \\ \hline
	45 & Random Forest            & \faDotCircleO & \faCheckCircle & \faCheckCircle & \faCheckCircle \\ \hline
	46 & Ridge                    & \faDotCircleO & \faCheckCircle & \faCheckCircle & \faCheckCircle \\ \hline
	47 & SAGA                     & \faDotCircleO & \faCheckCircle & \faCheckCircle & \faCheckCircle \\ \hline
	48 & SAG                      & \faDotCircleO & \faCheckCircle & \faCheckCircle & \faCheckCircle \\ \hline
	49 & SDCA                     & \faDotCircleO & \faCheckCircle & \faCheckCircle & \faCheckCircle \\ \hline
	50 & SGD                      & \faDotCircleO & \faTimesCircle & \faTimesCircle & \faTimesCircle \\ \hline
	51 & SVR                      & \faBullseye   & \faCheckCircle & \faCheckCircle & \faCheckCircle \\ \hline
	52 & SVRG                     & \faDotCircleO & \faTimesCircle & \faTimesCircle & \faTimesCircle \\ \hline
	53 & Stack. ST Hoeffding Tree & \faDotCircleO & \faCheckCircle & \faCheckCircle & \faCheckCircle \\ \hline
	54 & Theil-Sen                & \faBullseye   & \faCheckCircle & \faCheckCircle & \faCheckCircle \\ \hline
	55 & Transformed Target       & \faDotCircleO & \faCheckCircle & \faCheckCircle & \faCheckCircle \\ \hline
	56 & Tweedie                  & \faBullseye   & \faCheckCircle & \faCheckCircle & \faCheckCircle \\ \hline
	57 & XGBoost                  & \faBullseye   & \faCheckCircle & \faCheckCircle & \faCheckCircle \\ \hline
	58 & iSOUP Tree               & \faDotCircleO & \faCheckCircle & \faCheckCircle & \faCheckCircle \\ \hline
    \caption{List specifying the type of each machine learning algorithm and the experiments where they were used.
    Along with the table, \faDotCircleO~to indicate multi-output algorithms, and \faBullseye~to refer to single-output algorithms.
    \faCheckCircle~indicates whenever an algorithm was used with that specific experiment and \faTimesCircle~indicates otherwise.
    An algorithm was removed from the pipeline whenever it could not provide results given the computational infrastructure available or when it took longer than seven days for training over the dataset for each experiment ({\it i.e.}, 15-05, 15-25, and 30-50).}
\end{longtable}